\pdfoutput=1

\documentclass[11pt]{article}

\usepackage[]{acl}

\usepackage{times}
\usepackage{latexsym}

\usepackage[T1]{fontenc}

\usepackage[utf8]{inputenc}

\usepackage{microtype}

\usepackage{tikz}
\usepackage[inline]{enumitem}
\usepackage{amsmath}
\usepackage{cleveref}
\usepackage{svg}
\usepackage{booktabs}
\usepackage{subcaption}
\usepackage{svg}

\usepackage{supertabular}
\usepackage{algorithm}
\usepackage{algpseudocode}

\usepackage{circuitikz}
\usepackage{pifont}

\NewDocumentCommand{\rot}{O{45} O{1em} m}{\makebox[#2][l]{\rotatebox{#1}{#3}}}

\newcommand*{\EG}{e.g.}
\title{ 
Fact Recall, Heuristics or Pure Guesswork?\\
Precise Interpretations of Language Models for Fact Completion}

\def\authorsep{\hspace{0.3em}}

\author{Denitsa Saynova$^{1,2*}$ 
 \authorsep Lovisa Hagström$^{1,2*}$  \\ 
  \textbf{Moa Johansson}$^{1,2}$   
 \authorsep \textbf{Richard Johansson}$^{1,2}$
 \authorsep \textbf{Marco Kuhlmann}$^{3}$ \medskip\\
\null$^{1}$ Chalmers University of Technology \quad \null$^{2}$ University of Gothenburg
\quad \null$^{3}$ Linköping University\\
\texttt{\{saynova, lovhag\}@chalmers.se}}

\begin{document}
\maketitle\def\thefootnote{*}\footnotetext{Equal contribution.}\def\thefootnote{\arabic{footnote}}
\begin{abstract}

Language models (LMs) can make a correct prediction based on many possible signals in a prompt, not all corresponding to recall of factual associations. However, current interpretations of LMs fail to take this into account. For example, given the query ``Astrid Lindgren was born in'' with the corresponding completion ``Sweden'', no difference is made between whether the prediction was based on knowing where the author was born or assuming that a person with a Swedish-sounding name was born in Sweden. In this paper, we present a model-specific recipe -- \textsc{PrISM} -- for constructing datasets with examples of four different prediction scenarios: generic language modeling, guesswork, heuristics recall and exact fact recall. We apply two popular interpretability methods to the scenarios: causal tracing (CT) and information flow analysis. We find that both yield distinct results for each scenario. Results for exact fact recall and generic language modeling scenarios confirm previous conclusions about the importance of mid-range MLP sublayers for fact recall, while results for guesswork and heuristics indicate a critical role of late last token position MLP sublayers. 
In summary, we contribute resources for a more extensive and granular study of fact completion in LMs, together with analyses that provide a more nuanced understanding of how LMs process fact-related queries.
\end{abstract}

\section{Introduction}

Language models (LMs) trained on large corpora have been found to store significant amounts of factual information \citep{petroni-etal-2019-language}.  
While there are many research results documenting the fact proficiency of LMs \citep{pmlr-v202-kandpal23a,mallen-etal-2023-trust}, our understanding of how these models perform fact completion is still under development. Mechanistic interpretability is a growing area of research aiming to explain model behavior   \citep{elhage2021mathematical,geiger2021causal}, and has already yielded insights into where LMs store and process factual information \citep{meng2022locating,geva-etal-2023-dissecting,haviv-etal-2023-understanding}. 

\begin{figure}
    \centering
    \includegraphics[width=0.9\linewidth]{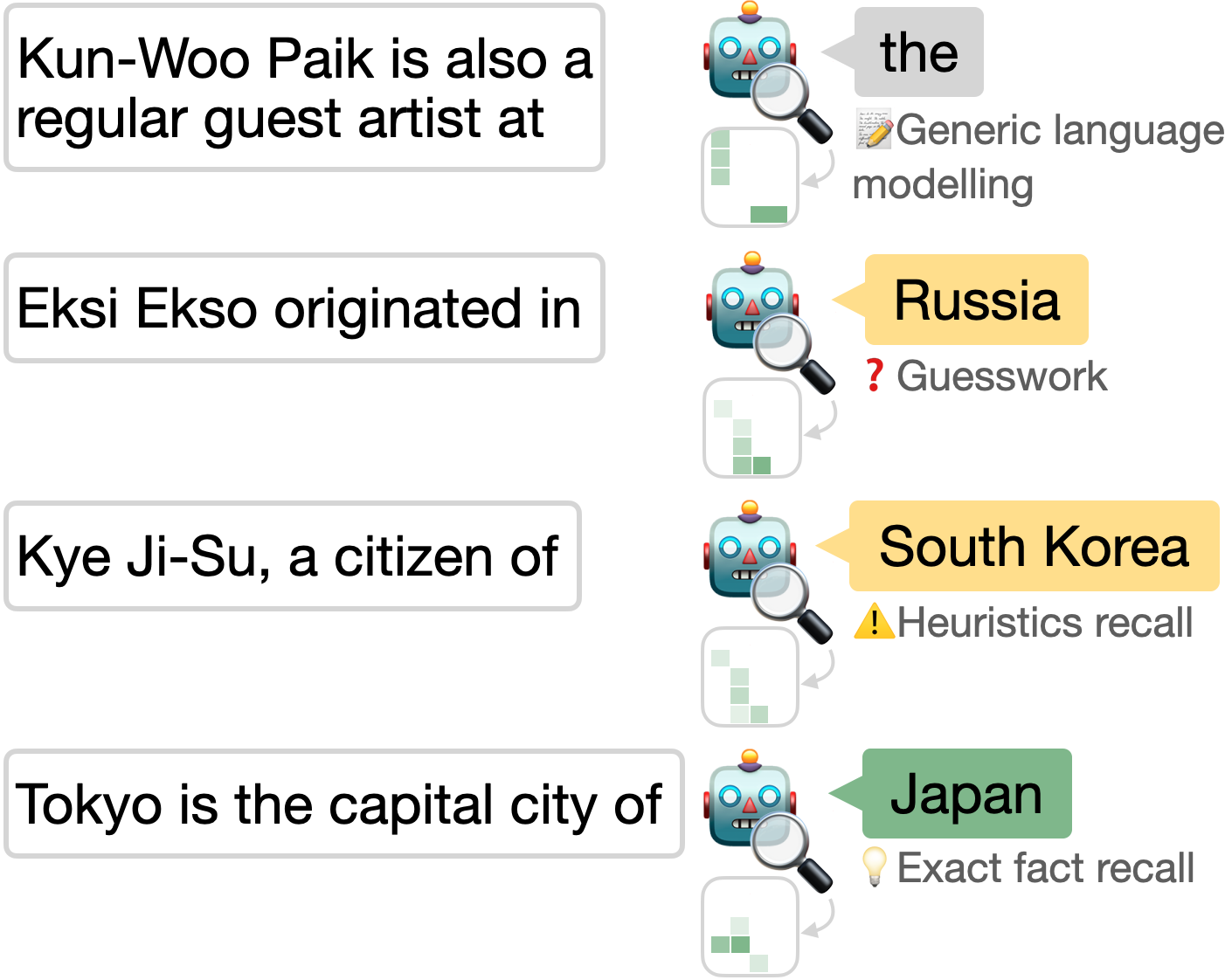}
    \caption{Prediction scenarios and corresponding prompt completion examples. Each scenario yields distinct interpretability results.} 
    \label{fig:mechanism_examples}
\end{figure}

While this body of work has broken new ground and provided us with interpretations of fact completion in LMs, it is limited to studies of \emph{only one type of scenario}. More specifically, in \citet{meng2022locating} and \citet{geva-etal-2023-dissecting}, the studies are limited to \emph{correct predictions}, and it is assumed that the model recalls facts for these. We hypothesize that this scenario in reality is a blend of multiple more fine-grained scenarios, as it is well known that LMs can make correct predictions based on many different signals in the prompt, not all corresponding to \emph{exact fact recall}. For example, LMs may pick up on spurious correlations to ``solve the dataset'' rather than the task exemplified by it \citep{zellers-etal-2019-hellaswag,niven-kao-2019-probing,mccoy-etal-2019-right}, and fact completion situations are no exception to this \citep{poerner-etal-2020-e,cao-etal-2021-knowledgeable,ladhak-etal-2023-pre}.
In bringing this distinction forward, we also ground our work in more formal studies of knowledge, where most scholars agree that guesswork should be accounted for and excluded in order to consider a model as having ``\textit{bona fide} knowledge''\cite{fierro-etal-2024-defining}.

In this work, we disentangle and detail four different \emph{prediction scenarios} for which LMs can be expected to show distinct behaviors (see \Cref{fig:mechanism_examples}). The scenarios are: 1)~\emph{Generic language modeling}, when the model does not respond with facts, such as when generating a story. 2)~\emph{Guesswork}, when the model responds with a fact but is uncertain. 3) \emph{Heuristics recall}, when the model uses shallow heuristics, \EG\ that people with Korean-sounding names are more likely to live in Korea. 4)~\emph{Exact fact recall}, when the model has indeed memorized the correct answer and recalls it for the prediction. We show how interpretations of LMs can be extended to these scenarios and how each scenario yields distinct interpretability results.

In particular, this work makes three main contributions. First, we propose a method, \textsc{PrISM}, for creating a diagnostic dataset with distinct test cases to enable more extensive and precise interpretations of fact completion in LMs (\S\ref{sec:recall-dataset}). We create and release \textsc{PrISM} datasets for GPT-2 XL, Llama 2 7B and Llama 2 13B, respectively. Second, our experiments with the interpretability method of causal tracing (CT) show that the models exhibit a complex behavior on the \textsc{PrISM} scenarios not captured by previous results (\S\ref{sec:CT-sensitivity-aggregation}). Third, our in-depth analysis of information flow confirms that models employ distinct inference mechanisms for the \textsc{PrISM} scenarios. For example, we observe contrasting results for exact fact recall compared to generic language modeling samples (\S\ref{sec:information-flow}). Taken together, our work expands on the scenarios that can be analyzed for interpretations of LMs for fact completion and yield new interpretations of LMs.\footnote{Dataset and code are available at \url{https://github.com/dsaynova/lm_interpretation_fact_completion}.}

\section{Background}\label{sec:related_work}
This section provides a brief background on fact completion and mechanistic interpretability topics relevant to our work.

\subsection{LMs and memorization}
A large body of interpretability work for fact completion situations is concerned with whether a model has memorized\footnote{Note that the type of memorization referred to here is of an abstracted representation of the fact rather than, as studied in some literature, of exact string memorization.} some fact or not, and where that fact is stored in the model parameters, also referred to as the \emph{parametric memory}. 
\citet{mallen-etal-2023-trust, pmlr-v202-kandpal23a} observe that queries asking for fact tuples rarely found in the LM training data are less likely to be known by the model. \citet{mallen-etal-2023-trust} take this one step further and use fact popularity (measured as Wikipedia page views) as a proxy for training data frequency to estimate the probability of a model knowing a fact. Conversely, \citet{liu2023prudent,basmov2024llms} use synthesized facts to simulate a training data frequency of 0 to study model behavior in the face of the unknown.

\subsection{LMs and heuristics}
Research into model performance on factual benchmarks has identified different factors affecting a model prediction. Accurate fact completions may stem from superficial cues and learned shallow heuristics, such as lexical overlap, person name bias\footnote{E.g., predicting \emph{Kye Ji-Su} to be a citizen of \emph{South Korea} due to the form of the name.} or prompt bias\footnote{E.g., predicting \emph{London} for ``Adam Doe was born in'' due to the training data showing strong correlations between \emph{London} and ``was born in'', disregarding the subject \emph{Adam Doe}.}
\cite{poerner-etal-2020-e, ladhak-etal-2023-pre,cao-etal-2021-knowledgeable}. 

While shallow heuristics are a natural by-product of the way LMs are trained and may provide a shortcut solution for some samples, they rely on disputable and overgeneralizing assumptions. 
Therefore, LMs relying on shallow heuristics in a fact prediction setting is generally undesirable \citep{mccoy-etal-2019-right}. For example, \citet{ladhak-etal-2023-pre} found that name bias leads to hallucinations and factually incorrect summaries by LMs. 

\subsection{LMs and random guesswork}
\citet{feng-etal-2024-dont} claim that reliable LMs need to be able to abstain from generating low-confidence outputs. Meanwhile, most open-source LMs are as of yet not equipped with the ability to abstain and, while there is a large body of research on this, there is no final verdict on the best method for measuring model confidence \citep{jiang-etal-2021-know, 8683359, burns2023discovering,yoshikawa-okazaki-2023-selective,zhao-etal-2024-knowing}.

\subsection{Interpretability and fact completion}

Recent work by \citet{meng2022locating, geva-etal-2023-dissecting,haviv-etal-2023-understanding} has focused on the inference process of LMs for fact completion for simple (subject, relation, object) fact tuples, such as subject \emph{Tokyo}, relation \emph{capital-of} and object \emph{Japan}, illustrated in \Cref{fig:mechanism_examples}. This body of work hypothesizes that LMs follow a distinct process when producing accurate fact completions. 
This hypothesis was originally posed by \citet{meng2022locating} based on aggregations of CT results, which revealed a decisive role of MLP modules at (last subject token, mid layer) positions for accurate fact completion predictions. This was reasoned to indicate that these modules \emph{recall fact associations} for a subject. 
Later results by \EG\ \citet{geva-etal-2023-dissecting} support the same conclusion. \footnote{
These knowledge localization efforts 
are somewhat orthogonal to work on model editing. While localization results from CT may not identify the optimal parameters for knowledge editing, this does not mean that CT is inaccurate for knowledge localization. It simply means that ``localization analysis might answer a different question than the question answered by model editing'' \citep{hase-does-localization-inform-editing}.}

\subsection{Causal tracing}
 
Causal tracing is a mechanistic interpretability method that has provided many interpretations of LMs \citep{stolfo-etal-2023-mechanistic,monea-etal-2024-glitch}. It was introduced by \citet{meng2022locating} and relies on the study of indirect causal effects.
By corrupting and restoring corrupted representations at different (token, layer) positions in a LM it is possible to infer what parts of the network are important for assigning a high probability to the predicted token with respect to the subject. The measured signal of model component importance is referred to as \emph{indirect effect}.

\citet{meng2022locating} also developed the \emph{CounterFact} dataset. Their conclusion is based on the 1,209 known samples from CounterFact for which GPT-2 XL is accurate. By now, it has been frequently used for the interpretation of LMs performing fact completion \citep{geva-etal-2023-dissecting}.

\subsection{Studies of information flow}

\citet{geva-etal-2023-dissecting} analyze \emph{information flow} in LMs for fact-related queries from CounterFact to understand how information is retrieved internally during inference. We mainly focus on their methods of \emph{attention knockout} to study from what (token position, layer) state critical information flows for the prediction and \emph{logit lens} to investigate attribute extraction in intermediate MLP and multi-head self-attention (MHSA) states. Using these methods, \citet{geva-etal-2023-dissecting} find three main ways in which information flows in LMs for fact-related predictions: 1) critical information flows from middle-upper subject position layers to the final token state corresponding to the prediction (``attribute extraction''), 2) critical information flows from early non-subject position layers (``relation propagation'') and 3) the subject position state is enriched with attributes by MLP layers before the attribute extraction takes place (``subject enrichment'').

\section{\textsc{PrISM} datasets for precise studies of prediction scenarios}\label{sec:recall-dataset}
We develop \textsc{PrISM} (Precise Identification of Scenarios for Model behavior) datasets to separate the different prediction scenarios in \Cref{fig:mechanism_examples}. This is motivated by an inspection of the 1,209 CounterFact samples which reveals 510 samples likely to rely on heuristics and 365 samples unlikely to have been memorized due to low popularity scores (see \Cref{app:knowns}). We argue that these issues make the CounterFact dataset unable to support precise and comprehensive interpretations of LMs, missing out on important distinctions and valuable insights into the workings of LMs. \textsc{PrISM} is developed to address these shortcomings.

\begin{table*}[htbp]%
    \small
    \centering%
    \begin{tabular}{llll}%
    \toprule%
     & GPT-2 XL & Llama 2 7B & Llama 2 13B \\%
    Scenario & \begin{tabular}[c]{@{}l@{}}\#samples (\#fact tuples)\end{tabular} & \begin{tabular}[c]{@{}l@{}}\#samples (\#fact tuples)\end{tabular} &  \begin{tabular}[c]{@{}l@{}}\#samples (\#fact tuples)\end{tabular}\\%
    \midrule%
    Generic LM & 1,000 (-) & 1,000 (-) & 1,000 (-) \\%
    Guesswork & 3,282 (3,181) & 2,917 (2,846) & 2,822 (2,220) \\%
    Heuristics & 8,352 (1,868) & 8,414 (1,960) & 9,224 (2,062)\\%
    Exact fact & 1,322 (191) & 5,481 (580) & 5,995 (601) \\%
    \bottomrule%
    \end{tabular}%
    \caption{Statistics for the \textsc{PrISM} datasets for each LM considered in our study.}%
    \label{tab:x-dataset}%
\end{table*}%

\begin{figure}
    \centering
    \includegraphics[width=\linewidth]{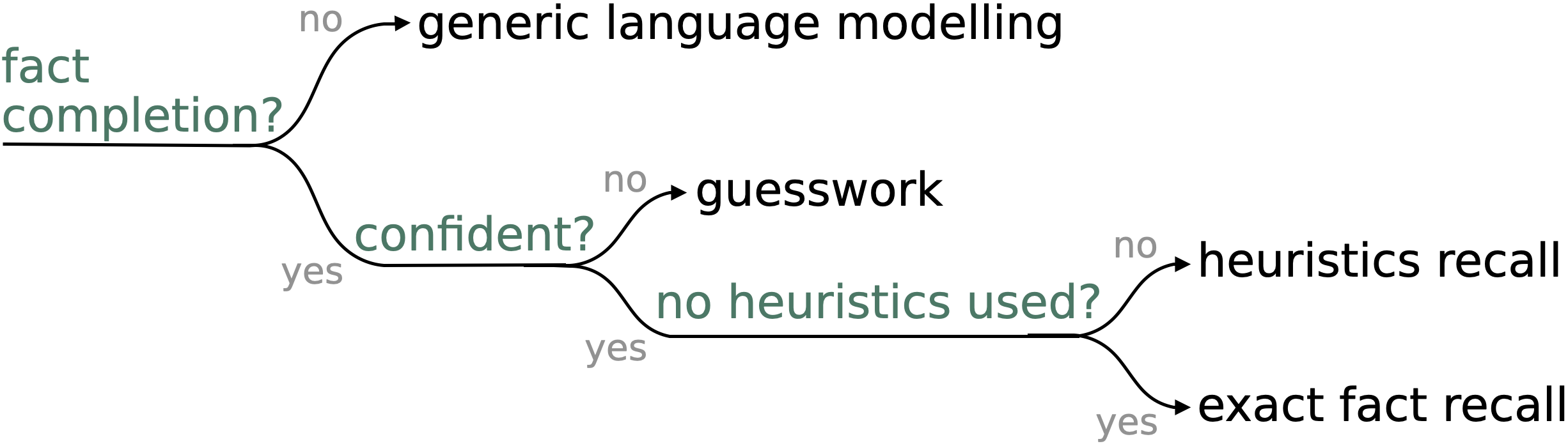}
    \caption{Diagnostic criteria (in green) for defining the four prediction scenarios (in black).}
    \label{fig:criteria-mechanisms}
\end{figure}

\textsc{PrISM} datasets are created by the identification and generation of samples corresponding to each of the four prediction scenarios in \Cref{fig:mechanism_examples}. The process relies on three diagnostic test criteria (\Cref{fig:criteria-mechanisms}).
Note that \textsc{PrISM} datasets are model-specific since they depend on model biases and parametric memories, which differ between LMs.

We develop \textsc{PrISM} datasets for GPT-2 XL \citep{radford2019language}, Llama 2 7B and Llama 2 13B. \citep{touvron2023llama}, respectively. General statistics for the datasets can be found in \Cref{tab:x-dataset} and examples corresponding to each prediction scenario can be found in \Cref{app:X-samples}. The samples and our subsequent analysis is focused on the English language. 
Further details on the implementation of the datasets can be found in \Cref{app:PrISM-creation}.
    
\subsection{Diagnostic criteria}
To create the \textsc{PrISM} datasets we propose three necessary and comprehensive diagnostic criteria for which we define measurements (see \Cref{fig:criteria-mechanisms}):
\begin{enumerate*}[label=(\arabic*)]
\item Does the prediction represent fact completion rather than generic language modeling?
\item Is the prediction confident and robust to insignificant signals in the prompt?
\item Is the prediction based on the exact factual information expressed in the query or on heuristics triggered by surface-level cues?
\end{enumerate*}
These criteria provide a more fine-grained testing setup compared to using a single accuracy-focused criterion, as in previous work.

\paragraph{Fact completion}\label{sec:framework-fact-completion}

Our first criterion is \emph{fact completion} -- whether a prompt and the model's prediction correspond to the setting of a model completing a fact. By making sure the model is studied only in fact completion situations, interpretations can be assumed to elucidate some fact processing behavior. This as opposed to, for example, the LM generating a story about unicorns, for which a different model behavior is assumed to take place.    
This criterion is already implicitly used in previous research on interpretations of LMs for fact completion \citep{petroni-etal-2019-language,meng2022locating,geva-etal-2023-dissecting}, we simply make it explicit. 

Following this body of work, for fact completion we limit ourselves to simple queries that express an incomplete fact (subject and relation), with the intent to let the LM generate the object as the next token.
We use ParaRel query templates to collect samples of fact completion \citep{elazar-etal-2021-measuring}. These are a set of rephrased expressions of a relation, where a different subject X can be substituted at the first position and an object Y is expected as the next token to be generated. For example, for the relation ``born-in'' we have templates such as ``[X] was born in [Y]'' or ``[X] is originally from [Y]'' (see \Cref{app:pararel-templates} for a full list). In total, 7 different relations are used, each with at least 5 different templates for query variations.

We define the measurement for fact completion as: If a query expresses an \emph{incomplete fact} (subject and relation) and the prediction corresponds to an entity or concept that has a \emph{valid type} (e.g. a place name when asking about location).
This excludes predictions such as ``the'', ``a'' and ``with''.

\paragraph{Confident prediction}\label{sec:framework-confidence}

Our second criterion is \emph{confident prediction} -- whether the prediction can be considered confident, e.g. by being robust across insignificant perturbations to the query.
Since most LMs by default cannot abstain from answering, we may end up in situations when a LM makes the correct prediction by chance while it randomly selects a token of the correct type (e.g. a city when asked for a birthplace) but has no relevant parametric knowledge of the specific fact. 

For the collection of \textsc{PrISM} samples, we opt for a definition of confidence grounded in desirable model behavior.
We proxy model confidence by consistency in the face of semantically equivalent queries \citep{elazar-etal-2021-measuring, portillo-wightman-etal-2023-strength,zhao-etal-2024-knowing} and measure it as the agreement across paraphrases from the ParaRel dataset \citep{elazar-etal-2021-measuring}.

We define the measurement for confident prediction as: If a prediction occurs among the \emph{top 3 predictions for at least 5 paraphrased queries}.
A prediction that only appears for one of the rephrased queries is deemed \emph{unconfident}. The thresholds were based on manual inspection to ensure adequate sampling of confident and unconfident samples. Since we use several tests to establish the type of prediction, we believe a slight variation of these threshold values will not lead to a substantial difference in the examples generated. Moreover, as observed in \Cref{sec:related_work}, there is no clear definition of \emph{confident prediction}, why we opted for creating our own. Other work may propose alternative definitions of confidence, \textsc{PrISM} can easily be adapted to these as well.

\paragraph{Usage of heuristics}\label{sec:framework-surface}
Our third and final criterion is \emph{no dependence on heuristics} -- indicating if the prediction is based on the exact factual information expressed in the prompt (subject and relation) rather than only on partial signals, i.e.\ heuristics. Predictions depending on heuristics indicate an over-reliance on unintended correlations in the training dataset based on surface forms of names or prompts, and are therefore unreliable \citep{cao-etal-2021-knowledgeable,mccoy-etal-2019-right,biran-etal-2024-hopping}. 

For the collection of \textsc{PrISM} samples we use two indicators: presence of surface-level cues and memorization estimation. If a model indicates it has learned a heuristic related to the prediction, it is likely this is used for completing the query. Additionally, if we know that the LM does not know the fact requested by a prompt but it still makes a confident prediction, we can assume that the prediction corresponds to some form of heuristics recall.

We use three types of filters for the detection of heuristics: Lexical overlap (between subject and prediction), person name bias and prompt bias filters. The two former filters are from \citet{poerner-etal-2020-e} and the latter is based on the findings by \citet{cao-etal-2021-knowledgeable} (\Cref{app:bias_filter}). We use fact popularity to estimate model knowledge, proxied by Wikipedia page views for the year 2019\footnote{Using the \href{https://wikitech.wikimedia.org/wiki/Analytics/AQS/Pageviews}{Pageview API}.}, based on work by \citet{mallen-etal-2023-trust}, where the authors find a strong correlation between popularity and memorization.

We define the measurement for usage of heuristics as: If a prediction is \emph{not} based on \emph{memorization} and the query contains \emph{surface-level cues}.

\subsection{Dataset creation}

Using the above criteria, we build \textsc{PrISM} datasets of (\emph{query}, \emph{prediction}) samples representative of each of four potential prediction scenarios: 1) generic language modeling, 2) random guesswork, 3) heuristics recall and 4) exact fact recall. In this section, we introduce definitions of the prediction scenarios based on our diagnostic criteria, illustrated in \Cref{fig:criteria-mechanisms}, and our method for producing the \textsc{PrISM} samples. Our goal is to create splits that contain a single prediction scenario rather than to identify all samples that correspond to that scenario. As such, we opt for stricter thresholds when needed to ensure high precision samples. Our approach is not intended to classify every instance into one of the four scenarios, but to produce high-quality examples of each one.

For the collection of \textsc{PrISM} samples, we consider the top 3 model predictions. By looking at multiple top tokens we break the over-reliance of previous work on accuracy of the top prediction and account for a larger portion of the LM output distribution.

\paragraph{Generic language modeling} We define samples corresponding to (\emph{fact completion: False}) as representative of a generic language modeling scenario. 

Generic language modeling samples are retrieved from Wikipedia.\footnote{We use \texttt{20220301.en} from HuggingFace at \url{https://huggingface.co/datasets/wikipedia}} We follow an approach similar to that of \citet{haviv-etal-2023-understanding} to ensure that we only collect samples corresponding to generic language modeling and not fact completion. The extraction is done by sampling sentences that start with the subject of the article in order to comply with the causal nature of the models and to allow for causal interventions on the subject. We discard the sentence if its natural continuation begins with a capital letter or a number (indicating this could be an entity and thus potentially fact completion).

\paragraph{Random guesswork} We define samples corresponding to (\emph{fact completion: True}, \emph{confident prediction: False}) as representative of a random guesswork scenario. 

For the collection of random guesswork samples, we first populate ParaRel templates with subjects and objects from LAMA \citep{petroni-etal-2019-language}. We retain (\emph{query}, \emph{prediction}) samples for which the prediction is a valid object from LAMA (e.g. a city when asked for birthplaces) but unconfident (i.e. it does not occur among the top 3 predictions for at least 5 paraphrased queries).

\paragraph{Heuristics recall} We define samples corresponding to (\emph{fact completion: True}, \emph{confident prediction: True}, \emph{no usage of heuristics: False}) as representative of a heuristics recall scenario. 

We collect these samples by populating ParaRel templates with synthetic fact tuples from a name generator (more details can be found in \Cref{app:biased-recall-creation}). Since the subjects are synthetic, facts about them cannot have been memorized by the model \citep{liu2023prudent,basmov2024llms}. Filtering confident predictions of valid objects for which a single type of bias is identified forms our heuristics recall samples.

\paragraph{Exact fact recall} We define samples corresponding to (\emph{fact completion: True}, \emph{confident prediction: True}, \emph{no usage of heuristics: True}) as representative of an exact fact recall scenario. The exact fact recall scenario corresponds to situations for which the LM can be expected to have memorized the full fact tuple expressed by the query and fetches this from memory for the prediction.

To obtain samples representative of exact fact recall, we again use the LAMA fact tuples. We collect predictions that are 1) confident, 2) not labeled as corresponding to any bias, 3) corresponding to a fact memorized by the LM, and 4) correct. 

\section{Interpretability and \textsc{PrISM}}\label{sec:CT-sensitivity}

We apply two mechanistic interpretability approaches -- CT and information flow analysis -- to test their sensitivity to each prediction scenario in \textsc{PrISM}.\footnote{We use the same hyperparameters as the original studies.} This allows us to evaluate the validity of previous interpretability results.

\begin{figure*}[h]
    \centering
    \begin{subfigure}[t]{0.369\textwidth}
        \centering
        \includegraphics[width=\linewidth]{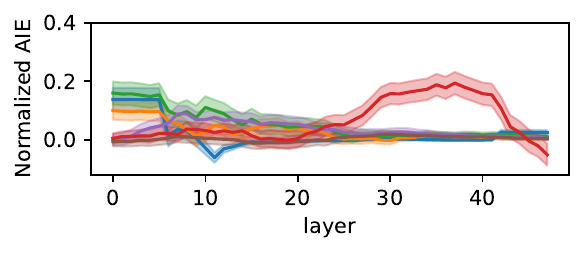}
        \vspace{-0.7cm}
        \caption{Generic language modeling samples.}
        \label{fig:CT_gpt2_generic_norm_lineplot1}
        
    \end{subfigure}%
    \begin{subfigure}[t]{0.3155\textwidth}
        \centering
        \includegraphics[width=\linewidth,trim={40pt 0pt 0pt 0pt},clip]{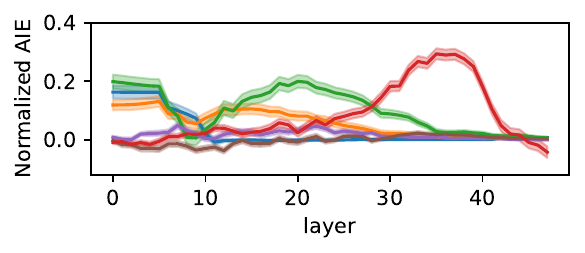}
        \vspace{-0.7cm}
        \caption{Guesswork samples.}
        \label{fig:CT_gpt2_guesswork_norm_lineplot}
    \end{subfigure}%
    \begin{subfigure}[t]{0.3155\textwidth}
        \centering
        \vspace{-2.35cm}
        \hspace{-2cm}
        \includegraphics[width=0.6\linewidth,trim={2pt 7pt 2pt 2pt},clip]{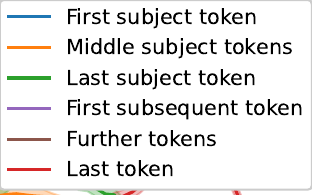}
    \end{subfigure}
    \begin{subfigure}[t]{0.369\textwidth}
        \centering
        \includegraphics[width=\linewidth]{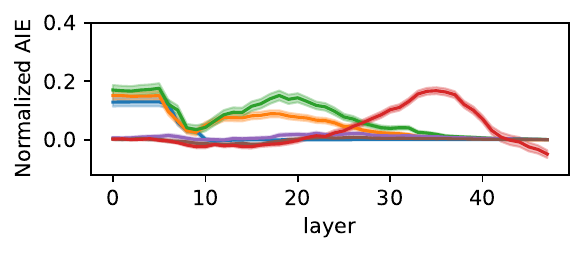}
        \vspace{-0.7cm}
        \caption{Heuristics recall samples.}
        \label{fig:CT_gpt2_biased_recall_norm_lineplot}
    \end{subfigure}%
    \begin{subfigure}[t]{0.3155\textwidth}
        \centering
        \includegraphics[width=\linewidth,trim={40pt 0pt 0pt 0pt},clip]{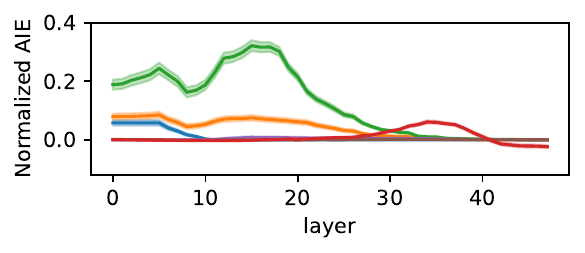}
        \vspace{-0.7cm}
        \caption{Exact fact recall samples.}
        \label{fig:CT_gpt2_exact_recall_norm_lineplot}
        
    \end{subfigure}%
    \begin{subfigure}[t]{0.3155\textwidth}
        \centering
        \includegraphics[width=\linewidth,trim={40pt 0pt 0pt 0pt},clip]{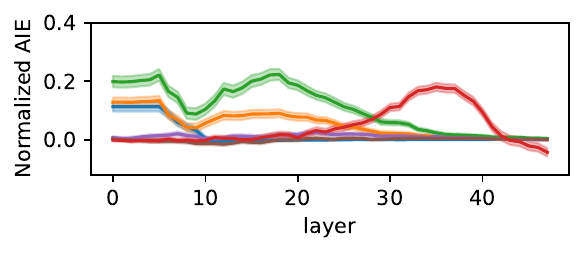}
        \vspace{-0.7cm}
        \caption{Combined samples.}
        \label{fig:CT_comb_lineplot}
    \end{subfigure}

    \caption{CT results for \textsc{PrISM} GPT-2 XL data. 1000 samples for each scenario in isolation. As well as 1000 combined samples (330 exact fact recall, 340 heuristics recall, 330 guesswork). Shaded regions indicate 95\% confidence intervals.}
    \label{fig:CT_comb_lineplots_gpt2}
\end{figure*}

\subsection{Causal tracing}\label{sec:CT-sensitivity-aggregation}

First, we investigate the sensitivity of CT to the \textsc{PrISM} scenarios. We aim to address the question \emph{Do CT results and the corresponding conclusions change with the underlying prediction scenario(s)?} 
CT measures the importance of different (token, layer) positions for a certain prediction. In line with \citet{meng2022locating}, we analyze the averaged indirect effects (AIE) per (token, layer) position, binning the input tokens into the following categories: first, middle, and last subject token; first subsequent token; further tokens; last token. 

To adjust for differences in absolute values of the probability of the predicted token, we take inspiration from \citet{hase-does-localization-inform-editing} and normalize results by how much the output token probability was reduced when corrupting the subject information. This measures the percentage of recoverable probability that was restored by patching a model state. For a detailed discussion of the effects of normalization, see \Cref{app:normalization}.

\Cref{fig:CT_comb_lineplots_gpt2} shows averaged normalized indirect effects of model states in GPT-2 XL for 1000 samples corresponding to each prediction scenario of \textsc{PrISM} in isolation as well as a combined plot of the 3 fact completion cases (exact fact recall, heuristics recall, and guesswork). 
The corresponding results for Llama 2 7B and Llama 2 13B can be found in \Cref{app:CT-comb-llama2} -- they support the same conclusions as reached for GPT-2 XL.

\paragraph{Prediction scenarios in isolation}
Results for the generic language modeling samples in \Cref{fig:CT_gpt2_generic_norm_lineplot1} indicate no critical role of last subject token position MLP states (used to indicate memory access). This further supports the original hypothesis that mid-layer MLP states act as memory storage, since they do not engage for samples that do not require memorization. 

We also observe how last token states in late layers are decisive for generic language modeling and guesswork. \citet{meng2022locating} recorded a similar peak in their results, while there is no clear hypothesis as to what information is processed here. We also note that the magnitude of the peak is much smaller for exact fact recall, indicating that this peak and the computations it corresponds to may signify a lack of exact fact recall. 

The results for the heuristics recall samples in \Cref{fig:CT_gpt2_biased_recall_norm_lineplot} show no decisive role of any particular token position state, while we note that it corresponds to a higher importance of last subject token states compared to generic language modeling, indicating that some memorized information is in use.

Results for exact fact recall samples (\Cref{fig:CT_gpt2_exact_recall_norm_lineplot}) are fundamentally different from those of the other isolated scenarios. The exact fact recall results show a clear peak in AIE in (last subject token, mid layer) MLP states. 
This is \emph{the only prediction scenario that clearly supports the same conclusion as previous work} in that (last subject token, mid layer) MLP states are decisive. This provides additional support for the hypothesis proposed by \citet{meng2022locating}, as the exact fact recall samples further emphasize the pattern interpreted to indicate the memory storage role of mid MLP layers.

\begin{figure*}[h!]
    \centering
    \begin{subfigure}[b]{0.2725\textwidth}
        \centering
        \includegraphics[width=\linewidth,trim={0pt 0pt 5pt 0pt},clip]{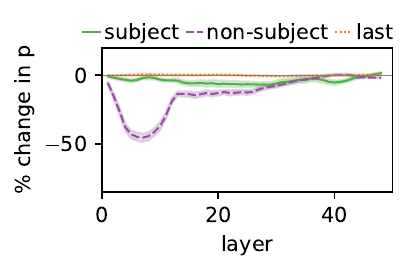}
        \vspace{-0.7cm}
        \caption{Generic language modeling}
        \label{fig:flow-generic}
    \end{subfigure}%
    \begin{subfigure}[b]{0.2425\textwidth}
        \centering
        \includegraphics[width=\linewidth,trim={22pt 0pt 0pt 0pt},clip]{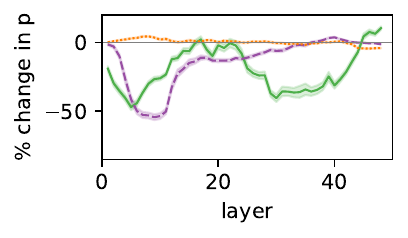}
        \vspace{-0.7cm}
        \caption{Guesswork}
        \label{fig:flow-guesswork}
    \end{subfigure}%
    \begin{subfigure}[b]{0.2425\textwidth}
        \centering
        \includegraphics[width=\linewidth,trim={22pt 0pt 0pt 0pt},clip]{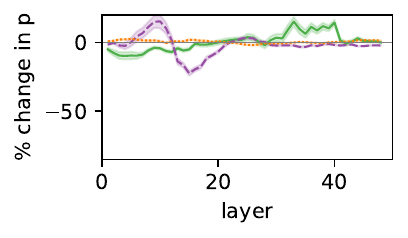}
        \vspace{-0.7cm}
        \caption{Heuristics recall}
        \label{fig:flow-heuristics}
    \end{subfigure}%
    \begin{subfigure}[b]{0.2425\textwidth}
        \centering
        \includegraphics[width=\linewidth,trim={22pt 0pt 0pt 0pt},clip]{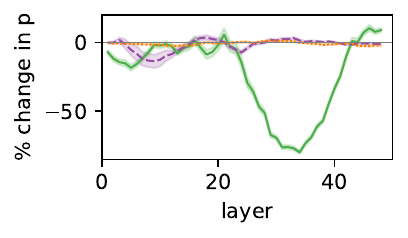}
        \vspace{-0.7cm}
        \caption{Exact fact recall}
        \label{fig:flow-exact-fact}
    \end{subfigure}
    \caption{Relative change in the prediction probability when intervening on attention edges to the last position for window sizes of 9 layers in GPT-2 XL on \textsc{PrISM} data. Shaded regions indicate 95\% confidence intervals.}
    \label{fig:flow}
\end{figure*}

\begin{figure*}[h!]
    \centering
    \begin{subfigure}[b]{0.2719\textwidth}
        \centering
        \includegraphics[width=\linewidth]{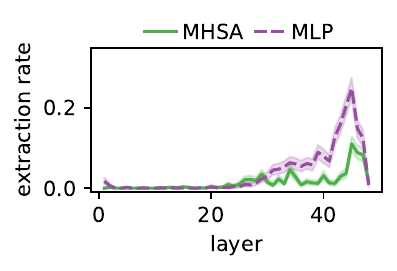}
        \vspace{-0.7cm}
        \caption{Generic language modeling}
        \label{fig:extraction-generic}
    \end{subfigure}%
    \begin{subfigure}[b]{0.2427\textwidth}
        \centering
        \includegraphics[width=\linewidth,trim={21pt 0pt 0pt 0pt},clip]{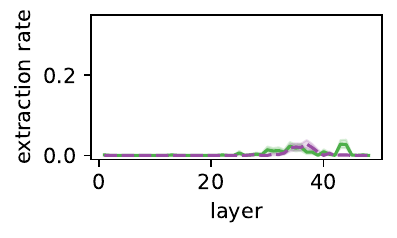}
        \vspace{-0.7cm}
        \caption{Guesswork}
        \label{fig:extraction-guesswork}
    \end{subfigure}%
    \begin{subfigure}[b]{0.2427\textwidth}
        \centering
        \includegraphics[width=\linewidth,trim={21pt 0pt 0pt 0pt},clip]{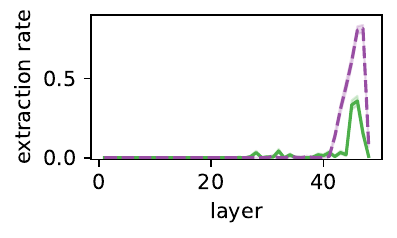}
        \vspace{-0.7cm}
        \caption{Heuristics recall}
        \label{fig:extraction-heuristics}
    \end{subfigure}%
    \begin{subfigure}[b]{0.2427\textwidth}
        \centering
        \includegraphics[width=\linewidth,trim={21pt 0pt 0pt 0pt},clip]{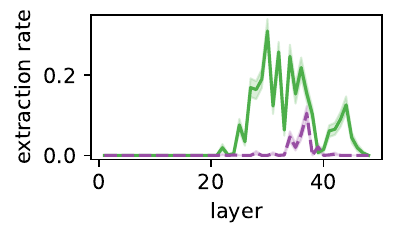}
        \vspace{-0.7cm}
        \caption{Exact fact recall}
        \label{fig:extraction-exact-fact}
    \end{subfigure}
    \caption{Attribute extraction rates across layers measured for \textsc{PrISM} GPT-2 XL data. MHSA and MLP indicate attribution extraction rates (top $k=1$) for multi-head self-attention states and multilayer perceptron states, respectively. Note the change in upper y axis limit from 0.35 to 0.9 for heuristics recall in (c). Shaded regions indicate 95\% confidence intervals.}
    \label{fig:extraction}
\end{figure*}

\paragraph{Aggregations of prediction scenarios}
To test the effects of analyzing mixed samples, we produce results for a mixture of fact completion scenarios. The combined plot of exact fact recall, heuristics recall, and guesswork samples in \Cref{fig:CT_comb_lineplot} generally reproduces the same CT results as observed in previous work, and thereby supports the same conclusion, i.e. (last subject token, mid layer) MLP states are decisive \citep{meng2022locating}. This indicates that model interpretations over samples mixing prediction scenarios are misleading as they may be dominated by the characteristics of the exact fact recall scenario. Potentially, this could be due to the exact fact recall samples generally corresponding to higher prediction confidences.

This supports our hypothesis that previous interpretability results may have been recorded over mixtures of prediction scenarios, as we can reproduce their results with a mixture. We do not use this as proof that these results are based on mixtures (see \Cref{app:knowns}). We rather wish to show how, without precise testing data, we can reach conclusions that are not supported by a large part of the studied sample (67\%).

\paragraph{Predictive potential}
We also investigate the potential for developing predictive systems based on our taxonomy and internal model states. 
We train a single-layer neural network with 50 neurons on a balanced four-way classification task. CT outputs AIE for each (token, layer) position, thus for each token, we obtain a vector of AIE values (for all model layers), similar to the ones in \Cref{fig:CT_comb_lineplots_gpt2}. Models are trained to predict the scenario based on the AIE vectors for the first and last subject token as well as the last prompt token (found to be most distinct between scenarios). The models achieve an accuracy of 0.72 for GPT-2 XL, 0.78 for Llama 2 7B, and 0.74 for Llama 2 13B (\Cref{app:ct_class}). We take this to indicate high correlations between internal model states and prediction scenarios, illustrating a potential of developing novel methods for disambiguating types of model behavior. 

\subsection{Information flow}\label{sec:information-flow}

In our second experiment we investigate the sensitivity of information flow analysis to the \textsc{PrISM} scenarios. Specifically, we leverage information flow localization via attention knockout and attribute extraction via logit lens \citep{geva-etal-2023-dissecting}. 

\Cref{fig:flow} shows the results from the attention knockout experiments and \Cref{fig:extraction} the attribute extraction results. Similarly to the CT results in \Cref{sec:CT-sensitivity-aggregation}, each prediction scenario corresponds to a unique information flow and extraction rate pattern. This further supports our claim that \textsc{PrISM} samples can be used to study LM behaviors for different prediction scenarios. A deeper analysis of the results is structured around two main questions:

\paragraph{Do our results support the same conclusions as reached in previous work?}
We focus on the two conclusions from \citet{geva-etal-2023-dissecting} related to 1) critical information flows from middle-upper subject position layers corresponding to a prediction probability decrease of 60\% and 2) critical information flows from early non-subject position layers corresponding to a probability decrease of 45\%. 

The results for the exact fact recall samples in \Cref{fig:flow-exact-fact} support conclusion (1). We even measure a stronger impact of subject attention connections (80\%) and higher extraction rates in \Cref{fig:extraction-exact-fact} compared to previous results. We also measure a significant impact of subject attention connections for guesswork samples (40\%), while the generic language modeling samples show very little impact of any subject position layers (7\%), further supporting conclusion (1). However, conclusion (2) is not supported by the results on the exact fact recall samples as information from early non-subject position layers is less important for exact fact recall (15\% to 45\%). High importance of non-subject positions information is instead observed for the guesswork samples (50\%) and generic language modeling samples (50\%). This indicates that the conclusions reached by \citet{geva-etal-2023-dissecting} largely are valid, while they seemingly average across guesswork and exact fact recall scenarios.

\paragraph{What inference process takes place for each prediction scenario?}
For the exact fact recall scenario, we have already noted how the inference process heavily relies on information from subject position layers and less on relation (non-subject) information. This is also supported by the clear importance of mid layer last subject position MLP states in \Cref{fig:CT_gpt2_exact_recall_norm_lineplot}. 

For the generic language modeling scenario, we note that only information from early non-subject position layers is critical for the final prediction in \Cref{fig:flow-generic}. We also record high extraction rates from the late \emph{MLP states} in \Cref{fig:extraction-generic} and a clear importance of late MLP states in \Cref{fig:CT_gpt2_generic_norm_lineplot1}. 
Based on these results, we propose that generic language modeling predictions mainly stem from late MLP computations solely based on the preceding tokens, with little notion of the subject under consideration.

For the guesswork scenario, we measure a flow quite similar to what was measured by \citet{geva-etal-2023-dissecting} (\Cref{fig:flow-guesswork}), with the exception of the early subject and non-subject position layer states being more important than late subject position states for guesswork. The extraction rates in \Cref{fig:extraction-guesswork} are much lower compared to those measured for the CounterFact and exact fact recall samples, however. A deeper analysis of the extraction rates including the top $k=3$ and top $k=10$ prediction extraction rates in \Cref{app:extraction-results} reveal that the predicted attribute seemingly is extracted, while it is not among the top-3 predictions in the model. Taken together with the results in \Cref{fig:CT_gpt2_guesswork_norm_lineplot}, we hypothesize that no top-3 prediction \emph{extraction} takes place, instead the probability of the final prediction is raised in the model via late last position MLP layer computations based on relation and subject information. While this conclusion potentially is not surprising, it confirms the quality of the generic modeling scenario and shows how it can be used as a baseline for \emph{non-fact recall behavior}.

For the heuristics recall scenario, we observe distinct patterns for both the information flow in \Cref{fig:flow-heuristics} and extraction rates in \Cref{fig:extraction-heuristics}. We reason that as the model has no notion of the given subject, little to no information can be extracted from ``memory'' in the subject position MLP layers, thus no probability drop from cutting attention to subject tokens. Therefore, the most critical information is transferred for early-mid non-relation position layers (20\%). Seemingly, the final prediction is extracted from the MLP layers at the last token position, leveraging some previous information and information about the latest token. A more fine-grained study that separates the heuristics making up the heuristics recall scenarios (prompt bias and person name bias) can be found in \Cref{app:dissection-results-heuristics}.

\section{Conclusion}

We identify four prediction scenarios that are fundamentally different and of differing reliability. These are \emph{exact fact recall}, \emph{heuristics recall}, \emph{guesswork} and \emph{generic language modeling}. We show that previous interpretability work for fact completion situations treat many of these as equivalent by using accuracy as the sole criterion for differentiating between different types of inference processes.
To facilitate precise interpretations of LMs, we present a method for creating \textsc{PrISM} datasets with samples that represent each of our identified prediction scenarios. We create \textsc{PrISM} datasets for GPT-2 XL, Llama 2 7B and Llama 2 13B, and use them to test the sensitivity of two influential interpretability methods, causal tracing and information flow analysis, to prediction scenario. We find that different prediction scenarios yield distinct interpretability results if studied in isolation. 
Taken together, our paper expands on and delineates fact completion scenarios for which we can interpret LMs. Our results highlight the importance of studying these scenarios in isolation and provide nuanced insights with respect to how LMs process information in fact completion situations. 

\section*{Limitations}
Similarly to previous interpretability work, our results are limited to auto-regressive models and subject-first template queries. Using the methods described in this paper, \textsc{PrISM} datasets can be constructed for other types of LMs, such as encoder-based models, while we leave this for future work. 

Moreover, the heuristics filters used for our dataset creation can only reveal the \emph{possibility} of shallow heuristics being used by the LM. We also observe some questionable samples that go undetected by the filters, indicating that the filters are leaky. Furthermore, we find signs of name-based heuristics for non-person subjects for which we have no applicable filters. The detection of these cases would rely on more advanced detection methods and is left for future work. By complementing our dataset creation with knowledge estimations and sampling of synthetic fact tuples, we should avoid most filter failures, while we cannot completely rule out the possibility of there being some problematic samples with \textsc{PrISM}. 

Even though we partition the \textsc{PrISM} samples based on whether the prediction is confident, we find that our results are sensitive to whether we investigate predictions with high or low probabilities from each partition. This indicates room for improvement for our method of detecting confident predictions, for which we already have noted a lack of comprehensive studies of model confidence metrics. Alternatively, this could be indicating a more fundamental issue with a qualitative difference in how models behave in low and high probability cases.

\section*{Ethical considerations}
Interpretability methods for fact completion situations are not directly associated with any ethical concerns. Neither is the LAMA dataset or synthetic fact tuples used in this work.

\section*{Acknowledgements}
A special thanks to Nicolas Audinet de Pieuchon as well as the research members of the CopeNLU group for their valuable feedback that helped us shape the paper. We would also like to thank the anonymous reviewers for their feedback and time.

This work was partially supported by the Wallenberg AI, Autonomous Systems and Software Program (WASP) funded by the Knut and Alice Wallenberg Foundation, by the Wallenberg AI, Autonomous Systems and Software Program -- Humanities and Society (WASP-HS) funded by the Marianne and Marcus Wallenberg Foundation and the Marcus and Amalia Wallenberg Foundation, and by the Swedish Excellence Center for Computational Social Science (SweCSS) funded by the Swedish Research Council through grant agreement no.\ 2022-06611. The computations were enabled by resources provided by the National Academic Infrastructure for Supercomputing in Sweden (NAISS) at Alvis partially funded by the Swedish Research Council through grant agreement no.\ 2022-06725.

\bibliography{anthology,custom}
\bibliographystyle{acl_natbib}

\clearpage

\appendix

\section{Computational resources}\label{app:comp_resources}
Experiments in this work are done on T4, A40 and A100 NVIDIA GPUs. Models used are GPT-2 XL (1.5B parameters), Llama 2 7B (7B parameters), and Llama 2 13B (13B parameters).

\section{Selection process of LAMA relations}\label{app:data}

The LAMA relations included in our \textsc{PrISM} dataset have been selected based on the following criteria:
\begin{enumerate}
    \item We only include relations that have multiple templates for which 1) the object comes last in order to fit the autoregressive setting and 2) the subject comes first in order to simplify the causal reasoning of intervening on the subject;
    \item  We exclude relations with a lot of overlap between the subject and object and relations for which the answers are highly imbalanced toward only a few alternatives.
\end{enumerate}
This corresponds to the relations P19 \emph{place of birth}, P20 \emph{place of death}, P27 \emph{country of citizenship}, P101 \emph{field of work}, P495 \emph{country of origin}, P740 \emph{location of formation} and P1376 \emph{capital of}.

\section{ParaRel templates}\label{app:pararel-templates}

We use the templates as described in \Cref{tab:pararel-templates} for the creation of \textsc{PrISM} queries.

\bottomcaption{\label{tab:pararel-templates}ParaRel templates used for all LAMA relations in our dataset creation.}
\begin{supertabular}[ht]{l l}
        \toprule
        Relation & Template \\
        \midrule
        P19 &   [X] was born in [Y] \\
            &   [X] is originally from [Y] \\
            &   [X] was originally from [Y] \\
            &   [X] originated from [Y] \\
            &   [X] originates from [Y] \\
            \midrule
        P20 &   [X] died in [Y] \\
            &   [X] died at [Y] \\
            &   [X] passed away in [Y] \\
            &   [X] passed away at [Y] \\
            &   [X] expired at [Y] \\
            &   [X] lost their life at [Y] \\
            &   [X]'s life ended in [Y] \\
            &   [X] succumbed at [Y] \\
            \midrule
        P27 &   [X] is a citizen of [Y] \\
            &   [X], a citizen of [Y] \\
            &   [X], who is a citizen of [Y] \\
            &   [X] holds a citizenship of [Y] \\
            &   [X] has a citizenship of [Y] \\
            &   [X], who holds a citizenship of [Y] \\
            &   [X], who has a citizenship of [Y] \\
            \midrule
        P101 &  [X] works in the field of [Y] \\
             &  [X] specializes in [Y] \\
             &  The expertise of [X] is [Y] \\
             &  The domain of activity of [X] is [Y] \\
             &  The domain of work of [X] is [Y] \\
             &  [X]'s area of work is [Y] \\
             &  [X]'s domain of work is [Y] \\
             &  [X]'s domain of activity is [Y] \\
             &  [X]'s expertise is [Y] \\
             &  [X] works in the area of [Y] \\
             \midrule
        P495 &  [X] was created in [Y] \\
             &  [X], that was created in [Y] \\
             &  [X], created in [Y] \\
             &  [X], that originated in [Y] \\
             &  [X] originated in [Y] \\
             &  [X] formed in [Y] \\
             &  [X] was formed in [Y] \\
             &  [X], that was formed in [Y] \\
             &  [X] was formulated in [Y] \\
             &  [X], formulated in [Y] \\
             &  [X], that was formulated in [Y] \\
             &  [X] was from [Y] \\
             &  [X], from [Y] \\
             &  [X], that was developed in [Y] \\
             &  [X] was developed in [Y] \\
             &  [X], developed in [Y] \\
             \midrule
        P740    & [X] was founded in [Y] \\
                & [X], founded in [Y] \\
                & [X] that was founded in [Y] \\
                & [X], that was started in [Y] \\
                & [X] started in [Y] \\
                & [X] was started in [Y] \\
                & [X], that was created in [Y] \\
                & [X], created in [Y] \\
                & [X] was created in [Y] \\
                & [X], that originated in [Y] \\
                & [X] originated in [Y] \\
                & [X] formed in [Y] \\
                & [X] was formed in [Y] \\
                & [X], that was formed in [Y] \\
                \midrule
        P1376   & [X] is the capital of [Y] \\
                & [X] is the capital city of [Y] \\
                & [X], the capital of [Y] \\
                & [X], the capital city of [Y] \\
                & [X], that is the capital of [Y] \\
                & [X], that is the capital city of [Y] \\
        \bottomrule
\end{supertabular}

\section{Creation process for \textsc{PrISM}}\label{app:PrISM-creation}

\subsection{Generic language modeling samples}\label{app:genericLM-creation}

Data is sampled from Wikipedia extraction \texttt{20220301.en} from HuggingFace at \url{https://huggingface.co/datasets/wikipedia}. This extraction contains around 6.5M pre-cleaned English Wikipedia articles. We perform the following steps:

\begin{enumerate}
    \item Select an entry (Wikipedia page) from the data. \emph{E.g.: the page for ``John Doyle (Irish artist)''}
    \item Select a single sentence from the page that begins with any part of the page title (i.e. it could be the surname, if the subject is a person). \emph{E.g.: ``Doyle continued to exhibit miniatures until 1835, but by then he was experiencing greater success with his political cartoons, printed using the new reproductive medium of lithography, beginning in 1827.''}
    \item We discard the sentence if it is: 1) shorter than 5 words, 2) with more than 3 capitalized words (likely to be section headings). \emph{E.g.:``Early life and family''}
    \item We cap the sentence at 10 words.\emph{E.g.: ``Doyle continued to exhibit miniatures until 1835, but by [then]''}
    \item We discard the sentence if its natural continuation begins with a capital or number (indicating this could be an entity and thus potentially fact completion). \emph{E.g.: ``Doyle won a gold medal in [1805].''}
\end{enumerate}
We repeat this until we have 1000 datapoints (for 1000 unique entries in the data). For CT experiments, we trace the next token, freely predicted by the model.  
\subsection{Guesswork samples}\label{app:guesswork-creation}
To get examples of guesswork, we follow a process as described below and apply it to all fact tuples from LAMA\footnote{\url{https://github.com/facebookresearch/LAMA}} corresponding to the relations P19 \emph{place of birth}, P20 \emph{place of death}, P27 \emph{country of citizenship}, P101 \emph{field of work}, P495 \emph{country of origin}, P740 \emph{location of formation} and P1376 \emph{capital of} and the corresponding ParaRel templates for these relations:

\begin{itemize}
    \item[] \textbf{for each} relation $r$ (e.g. ``P740'' \emph{location of formation}):
    \item[] \textbf{for each} LAMA subject $s$ for relation $r$ (e.g. ``Sonar Kollektiv''):
    \begin{itemize}
        \item[] \textbf{if} popularity score for $s$ > 1000 \textbf{then} discard all examples for the subject (likely to have been memorized)
        \item[] \textbf{else} create empty list $L$ for the tuple results
    \end{itemize}
    \item[] \textbf{for each} ParaRel template $t$ for relation $r$ (e.g. ``[X] originated in [Y]''):
    \begin{itemize}
        \item[] predictions $=$ top 3 predictions for $(s,t)$ (e.g. ``Sonar Kollektiv originated in'')
    \end{itemize}
    \item[] \textbf{for each} $p$ in predictions:
    \begin{itemize}
        \item[] \textbf{if} $p$ is trivial (e.g. ``the'') \textbf{then} discard tuple $(s,t,p)$
        \item[] \textbf{else} add $(s,t,p)$ to $L$
    \end{itemize}
    \item[] \textbf{end for each} $p$  
    \item[] \textbf{end for each} $t$  
    \item[] \textbf{if} count$(s,*,p)$ in $L$ $=$ 1 \textbf{then} add $(s,t,p)$ to guesswork sample (that is, if a particular answer $p$ occurs within the top 3 predictions for only one template it is considered guesswork) 
    \item[] \textbf{end for each} $s$ 
    \item[] \textbf{end for each} $r$ 
\end{itemize}
We measure popularity score proxied by Wikipedia page views for year 2019\footnote{Using the \href{https://wikitech.wikimedia.org/wiki/Analytics/AQS/Pageviews}{Pageview API}.} following \citep{mallen-etal-2023-trust}. We label a prediction as ``likely to be memorized'' if it corresponds to an average page view rate above 1000, as queries with lower popularity scores are unlikely to have been memorized \citep{mallen-etal-2023-trust}.

We determine if a prediction is trivial by only accepting a prediction as non-trivial if it's the correct answer for at least one data point in LAMA (for the same relation).

\subsection{Heuristics recall samples}\label{app:biased-recall-creation}

We create heuristics recall samples following a similar algorithm as for exact fact recall. Rather than starting from LAMA subject-object data points we produce a set of synthetic (non-existent) entities to populate the ParaRel templates to simulate a popularity score of 0 (not present in training data) and thus ensure no memorization. 

\subsubsection*{Sampling procedure}

We create the synthetic data by following the steps for each relation in (P19, P20, P27, P101, P495, P740 and P1376):
\begin{enumerate}\addtolength{\itemsep}{-0.5\baselineskip}
    \item Identify subject type distributions for the selected relations. \emph{E.g. For relations P19, P20, P27 and P101, based on the ``subject type constraint'' from Wikidata the only allowed subject type is person.}
    \item Generate subjects of the required types using \url{https://www.fantasynamegenerators.com}. \emph{E.g. For person subjects, we generate a mixture of DND human, Russian, French, German, Korean, and Japanese names}
    \item Perform de-duplication and check against Wikidata that no subject corresponds to a real entity.\footnote{The code for automatically querying Wikidata for real entities is provided as part of our code.} The Wikidata check is done on a label level, since the generated names are pure strings. This limits our ability to check for a subject's existence, as we can only find exact matches. 
\end{enumerate}
For the collected synthetic samples we apply the following algorithm:

\begin{itemize}
    \item[] \textbf{for each} relation $r$ (e.g. ``P19'' \emph{place of birth}):
    \item[] \textbf{for each} synthetic subject $s$ for relation $r$ (e.g. ``Serok Nuvrome''):
    \begin{itemize}
        \item[] create empty list $L$ for the tuple results
    \end{itemize}
    \item[] \textbf{for each} ParaRel template $t$ for relation $r$ (e.g. ``[X] was born in [Y]''):
    \begin{itemize}
        \item[] predictions $=$ top 3 predictions for $(s,t)$ (e.g. ``Serok Nuvrome was born in'')
    \end{itemize}
    \item[] \textbf{for each} $p$ in predictions:
    \begin{itemize}
        \item[] \textbf{if} $p$ is trivial (e.g. ``the'') \textbf{then} discard tuple $(s,t,p)$
        \item[] \textbf{if} $p$ is based on a single type of heuristics \textbf{then} add $(s,t,p)$ to $L$
        \item[] \textbf{else} discard tuple $(s,t,p)$
    \end{itemize}
    \item[] \textbf{end for each} $p$  
    \item[] \textbf{end for each} $t$  
    \item[] \textbf{if} count$(s,*,p)$ in $L$ $\geq$ 5 \textbf{then} add all $(s,*,p)$ tuples to heuristic recall sample (that is, if a particular answer $p$ occurs within the top 3 predictions for at least 5 templates it is considered confident) 
    \item[] \textbf{end for each} $s$ 
    \item[] \textbf{end for each} $r$ 
\end{itemize}
We identify if a prediction is trivial by the following set of rules: If the object type is a named entity (e.g. place names), we allow any generation beginning with a capital letter. This covers all relations apart from P101. For P101, we query Wikidata -- we check if there exist any (s,r,o) Wikidata entry where the object label matches the generated token.

We measure if a prediction is based on heuristics by applying 3 filters explained in more detail in \Cref{app:bias_filter}.

\subsubsection*{Distribution of generated subject types}

To approximate realistic data distributions, we generate subjects based on the LAMA subject types.

For relations P19, P20, P27, P101, only allowed subject is person, so we generate 1000 fantasy (Dungeons and Dragon human) names and 200 of each German, Korean, Russian, French, and Japanese synthetic names. For P1376 (capital of), we generate 100 of each Central Africa, Central America, Central Asia, East Asia, East Europe, Middle Eastern, West Europe sounding town names.

Relation P740 (location of formation) has a mixture of subject types (top 5 categories are shown in \Cref{tab:lama_p740}). Based on this, we generate 500 musical groups and 500 company names.

\begin{table}[h]
\centering
\begin{tabular}{ll}
\toprule
type                    & frequency \\
\midrule
musical group           & 505       \\
business                & 105       \\
public company          & 52        \\
rock band               & 29        \\
human                   & 20        \\
other                   & 225      \\
\bottomrule
\end{tabular}
\caption{\label{tab:lama_p740}Top 5 categories of subjects from LAMA for relation P740. The category ``other'' contains 102 different entity types with less than 15 instances each.}
\end{table}

Relation P495 (country of origin) has a diverse set of media and entertainment related subjects in LAMA (see \Cref{tab:p495_lama}). We generate 500 music groups and 100 of each anime and manga, book, newspaper, and magazine names.

\begin{table}[h]
\centering
\begin{tabular}{ll}
\toprule
type                & frequency \\
\midrule
television          & 175       \\
magazine            & 23        \\
music               & 145       \\
film                & 225       \\
anime, comic, manga & 48        \\
not found           & 65        \\
other               & 228      \\
\bottomrule
\end{tabular}
\caption{\label{tab:p495_lama}Top 5 categories of subjects from LAMA for relation P495. The category ``other'' contains 103 different entity types with less than 13 instances each.}
\end{table}

\subsubsection*{Analysis of the heuristics recall samples in \textsc{PrISM}}\label{app:X-biased-recall-analysis}
Our heuristics recall analysis identifies samples that are confident, but for which no bias is detected. This can be counter-intuitive, as we do not expect the model to be able to make confident prediction when it has no bias to guide it.  
These examples are excluded from the \textsc{PrISM} samples, but we perform a deeper analysis of the 1,771 cases from GPT-2 XL. 

6 instances identify the location of formation (P740) of ``Oasis of Prejudice'' as ``London'' (not identified as prompt bias, since the prompt bias check produces mostly years, indicating time to be the more natural interpretation of the queries). 9 instances from P101 (field of work) show the model potentially ignoring part of the query, by connecting ``Nina Schopenhauer'' with ``philosophy'' (potentially conflated with the philosopher Arthur Schopenhauer) and ``Roch Chagnon'' with ``anthropology'' (potentially conflated with the anthropologist Napoleon Chagnon). 23 examples of relation P495 show association of 5 fictional entities with Japan (3 of these contain the word ``Berserk'' -- a possible conflating pattern with the manga of the same name). Further 790 examples come from relations P19 (born in) and P27 (citizen of). Some of these could be examples of a stronger association overwriting the expressed tuple (e.g. ``Adolphe Trudeau'' associated with ``Quebec''), others may point to weaknesses of our name bias detection method. Finally, the most represented relation is P1376 (capital of) with 938 examples. This relation does not lend itself to our subject name bias filter, however, we suspect a linguistic correlation between city names and countries may exist and those surface level signals can potentially explain some of the predictions. 

This analysis confirms our concerns related to the coverage of the implemented heuristics recall filters. Evidently, there are some heuristics that go undetected by our filters. This is why we supplement the bias identification filters with memorization: For heuristic recall we simulate no memorization by using synthetic data and for exact fact recall we filter on high subject popularity (found to correlate well with memorization \citep{mallen-etal-2023-trust}).

\subsection{Exact fact recall samples}\label{app:exact-fact-recall-creation}
To get queries for which the LM performs exact fact recall, we follow a process as described below and apply it to all fact tuples from LAMA\footnote{\url{https://github.com/facebookresearch/LAMA}} corresponding to the relations P19 \emph{place of birth}, P20 \emph{place of death}, P27 \emph{country of citizenship}, P101 \emph{field of work}, P495 \emph{country of origin}, P740 \emph{location of formation} and P1376 \emph{capital of} and the corresponding ParaRel templates for these relations:

\begin{itemize}
    \item[] \textbf{for each} relation $r$ (e.g. ``P19'' \emph{place of birth}):
    \item[] \textbf{for each} LAMA subject $s$ for relation $r$ (e.g. ``Thomas Ong''):
    \begin{itemize}
        \item[] \textbf{if} popularity score for $s$ < 1000 \textbf{then} discard all examples for the subject
        \item[] \textbf{else} create empty list $L$ for the tuple results
    \end{itemize}
    \item[] \textbf{for each} ParaRel template $t$ for relation $r$ (e.g. ``[X] was born in [Y]''):
    \begin{itemize}
        \item[] predictions $=$ top 3 predictions for $(s,t)$ (e.g. ``Thomas Ong was born in'')
    \end{itemize}
    \item[] \textbf{for each} $p$ in predictions:
    \begin{itemize}
        \item[] \textbf{if} $p$ is based on heuristics \textbf{then} discard tuple $(s,t,p)$
        \item[] \textbf{if} $p$ is incorrect \textbf{then} discard tuple $(s,t,p)$ 
        \item[] \textbf{else} add $(s,t,p)$ to $L$
    \end{itemize}
    \item[] \textbf{end for each} $p$  
    \item[] \textbf{end for each} $t$  
    \item[] \textbf{if} count$(s,*,p)$ in $L$ $\geq$ 5 \textbf{then} add all $(s,*,p)$ tuples to exact fact recall sample (that is, if a particular answer $p$ occurs within the top 3 predictions for at least 5 templates it is considered confident) 
    \item[] \textbf{end for each} $s$ 
    \item[] \textbf{end for each} $r$ 
\end{itemize}
We measure popularity score proxied by Wikipedia page views for year 2019\footnote{Using the \href{https://wikitech.wikimedia.org/wiki/Analytics/AQS/Pageviews}{Pageview API}.} following \citep{mallen-etal-2023-trust}.

We measure if a prediction is based on heuristics by applying 3 filters explained in more detail in \Cref{app:bias_filter}.

We categorize the predictions into ``correct'' or ``incorrect'' using the LAMA gold labels. For Llama 2 models we categorize a prediction as ``correct'' if it has more than 3 characters and fully matches the start of the gold label. This is necessary since the tokenizers for these model are more prone to split the gold labels into small tokens.

\subsection{Detection filters for heuristics}\label{app:bias_filter}

Our detection of heuristics is based on 3 filters. 
\paragraph{Lexical overlap} Dependence on this heuristic is considered plausible if there is a string match between the prediction and the subject. \emph{E.g. ``San Salcos, the capital of [Sal]''} 
\paragraph{Name bias} We based this on model predictions for prompts expressing only a part of the requested fact. We query with the following prompts: ``[X] is a common name in the following city:'' and ``[X] is a common name in the following country:''. Where X is replaced with the subject name to check for bias. If any of the top 10 token predictions for these queries matches the model prediction for the full fact query, we mark that (\emph{query}, \emph{prediction}) pair as corresponding to person name bias.
We can detect person name bias for relations P19, P20, P27, used in \textsc{PrISM} and additionally for P103 and P1412, present in CounterFact.
\paragraph{Prompt bias} We use the original prompt templates as defined by ParaRel and replace the subject placeholder with generic substitutions. We use the substitutions described in \Cref{tab:prompt-bias-subject-subs} for each relation. We also remedy basic capitalization and grammar errors that might surface from this automated prompt creation. An example of a prompt for detecting prompt bias for ``Tokyo is the capital city of [Y]'' is ``It is the capital city of [Y]''. If the top prediction for the former query is found among the top 10 token predictions for the latter query, the former query and corresponding prediction is marked as based on prompt bias. 

\begin{table}[h]
    \centering
    \begin{tabular}{l l}
    \toprule
        Relation & Subject substitutions \\
    \midrule
        P19 & [He, She] \\
        P20 & [He, She] \\
        P27 & [He, She]  \\
        P101 & [He, She] \\
        P495 & [It] \\
        P740 & [It, The organisation] \\
        P1376 & [It, The city] \\
    \bottomrule
    \end{tabular}
    \caption{Subject substitutions used for constructing prompts to detect prompt bias.}
    \label{tab:prompt-bias-subject-subs}
\end{table}

\section{Examples from \textsc{PrISM}}\label{app:X-samples}
Here, we include a few examples to illustrate the content of \textsc{PrISM} for different prediction scenarios. See \Cref{tab:examples,tab:fact-recall-examples,tab:guesswork-samples,tab:heuristics-samples,tab:generic-samples}.

\begin{table*}[htbp]
    \begin{tabular}{llllccl}
    \toprule
    Scenario  & Prompt                                                         & Prediction & Gold label   & Conf & Pop & Bias   \\
    \midrule
    generic LM & Nara also enjoyed success in & the        & singles      & -          & -          & -      \\
    generic LM & Benjamin later joined a number of  & other        & clubs    & -          & -          & -      \\
    guesswork  & Sonar Kollektiv originated in                                  & Russia     & Berlin       & 1          & 215          & -      \\
    guesswork  & Joseph Clay was originally from                                & Ohio       & Philadelphia & 1          & 273          & -      \\
    heuristics & Serok Nuvrome, a citizen of & Ukraine      & -            & 6          & 0          & name   \\
    heuristics & Balo Windhair has a citizenship of                             & Canada     & -            & 5          & 0          & prompt \\
    exact fact & Thomas Ong is a citizen of                                     & Singapore  & Singapore    & 7          & 1418       & none   \\
    exact fact & Shibuya-kei, that was created in                               & Japan      & Japan        & 8          & 5933       & none  \\
    \bottomrule
    \end{tabular}
    \caption{Samples from \textsc{PrISM} for GPT-2 XL designed to trigger different prediction scenarios. Conf(idence) measures how often the prediction was made, pop(ularity) measures page view rate and bias indicates detected bias when applicable.}
    \label{tab:examples}
\end{table*}

\begin{table*}[h]
    \centering
    \begin{tabular}{l l l l l}
    \toprule
    Model & Query     & Prediction & Subject popularity & Gold label \\
    \midrule
    GPT-2 XL & Thomas Ong is a citizen of & Singapore     & 1418 & Singapore \\
    & Shibuya-kei, that was created in & Japan & 5933 & Japan \\
    & Palermo is the capital of & Sicily & 34273 & Sicily \\
    Llama 2 7B & Disco Biscuits was created in & Philadelphia & 3719 & Philadelphia \\
    & Don Broco, that was started in & Bed & 6984 & Bedford \\
    & Nikephoros III Botaneiates & Constantin & 1859 & Constantinople \\
    & \; passed away in & & & \\
    \bottomrule
    \end{tabular}
    \caption{(\emph{query}, \emph{prediction}) exact fact recall samples from \textsc{PrISM} for GPT-2 XL and Llama 2 7B.}
    \label{tab:fact-recall-examples}
\end{table*}

\begin{table*}[h]
    \centering
    \begin{tabular}{l l l l l}
    \toprule
    Model & Query     & Prediction & Rank & Gold label \\
    \midrule
    GPT-2 XL & Sonar Kollektiv originated in     & Russia & 1 & Berlin \\
    & Haydn Bendall is originally from & England & 2 & Essex \\
    & Joseph Clay was originally from & Ohio & 2 & Philadelphia \\
    Llama 2 7B & Jean Trembley originated from & France & 2 & Geneva \\
    & Dansez pentru tine, that originated in & France & 2 & Romania \\
    & Milton Wright is originally from & Chicago & 2 & Georgia \\
    \bottomrule
    \end{tabular}
    \caption{(\emph{query}, \emph{prediction}) random guesswork samples from \textsc{PrISM} for GPT-2 XL and Llama 2 7B.}
    \label{tab:guesswork-samples}
\end{table*}

\begin{table*}[h]
\centering
\begin{tabular}{llll}
\toprule
Model & Query & Prediction & Bias \\
\midrule
GPT-2 XL & Hirashima Hideyoshi, who has a citizenship of & Japan & name \\
 & Balo Windhair has a citizenship of & Canada & prompt \\
 & Olre Hellspirit was originally from & Hell & lexical \\
Llama 2 7B & Ha Songmin, who has a citizenship of & South (Korea) & name \\
 & Wanda Hagel holds a citizenship of & Canada & prompt \\
 & Limanaga, the capital city of & Lim & lexical \\
 \bottomrule
\end{tabular}
\caption{(\emph{query}, \emph{prediction}) heuristics recall samples from \textsc{PrISM} for GPT-2 XL and Llama 2 7B.}
\label{tab:heuristics-samples}
\end{table*}

\begin{table*}[h]
\centering
\begin{tabular}{llll}
\toprule
Model & Query & Prediction & Gold label \\
\midrule
GPT-2 XL & Dexmedetomidine is notable for its ability to provide sedation & and & without \\
 & Solomon also defended the network's choice of games to & air & broadcast \\
 & Walker added an immense amount of material to the & book & collections \\
Llama 2 7B & Dexmedetomidine is notable for its ability to provide sedation & and & without \\
 & Solomon also defended the network's choice of games to & air & broadcast \\
 & Walker added an immense amount of material to the & original & collections \\
 \bottomrule
\end{tabular}
\caption{(\emph{query}, \emph{prediction}) generic language samples from \textsc{PrISM} for GPT-2 XL and Llama 2 7B.}
\label{tab:generic-samples}
\end{table*}

\clearpage
\clearpage

\section{Inspection of CounterFact}\label{app:knowns}
In this section we assess the applicability of using the 1,209 known CounterFact samples for interpreting LMs in fact completion situations. First, we investigate what prediction scenarios are found for GPT-2 XL for the collection of the (\emph{query}, \emph{prediction}) samples (\Cref{sec:predmech-counter}). We find samples in the dataset likely to correspond to heuristics recall (510 samples) as opposed to exact fact recall (478 samples). Second, we inspect the total effects measured with the causal tracing approach for the dataset to find quality issues (\Cref{app:knowns-te-details}). Last, we observe a set of problematic samples with negated queries in CounterFact (\Cref{app:knowns-not}). Taken together, our results show that the dataset struggles to support precise and accurate interpretations of LMs. Our proposed \textsc{PrISM} dataset does not suffer from the aforementioned limitations.

\subsection{Prediction scenarios}\label{sec:predmech-counter}

We inspect the CounterFact dataset for three of four prediction scenarios. The baseline prediction scenario corresponding to generic language modeling is skipped for the analysis as the dataset should not trigger this scenario by virtue of its creation process.

\paragraph{Random guesswork}
We cannot detect samples in CounterFact corresponding to random guesswork as our implementation of the confidence criterion is incompatible with the dataset. The dataset only provides one prompt per fact, while we require multiple prompt variations per fact to estimate confidence. This does not mean that there are no samples in the CounterFact dataset corresponding to random guesswork, it only means that we are unable to detect them. As a result, some of the samples below identified to correspond to heuristics recall or exact fact recall may actually correspond to random guesswork, as we are unable to separate these samples beforehand.

\paragraph{Heuristics recall}
\begin{table*}[h]
    \centering
    \begin{tabular}{l l l}
        \toprule
        Query & Prediction & Bias type \\
        \midrule
        MacApp, a product created by & Apple & Prompt \\
        Giuseppe Angeli, who has a citizenship of & Italy & Person name \\
        The original language of La Fontaine's Fables is a mixture of & French & Prompt \\
        \bottomrule
    \end{tabular}
    \caption{Examples of queries and predictions from the known CounterFact dataset that potentially correspond to bias. The predictions and analysis has been performed for GPT-2 XL.}
    \label{tab:knowns-bias-examples}
\end{table*}

We check for predictions based on shallow heuristics for the known CounterFact samples produced using GPT-2 XL. We find a total of 510 samples that may correspond to heuristics recall, of which 335 samples correspond to prompt bias, 155 to name bias and 20 to both name and prompt bias.\footnote{There are a total of 205 samples corresponding to person names for which we can check for name bias, meaning that we detect name bias in 92.5\% of all cases.} No lexical overlap between sample subject and object is found. Some examples marked for bias can be found in \Cref{tab:knowns-bias-examples}. 

\begin{table}[h]
    \centering
    \begin{tabular}{l l}
        \toprule
        Popularity score & \# of samples \\
        \midrule
        $(0, 100]$ & 61 \\
        $(100, 1000]$ & 304 \\
        $(1000, 10000]$ & 379 \\
        $(10000, 1176235]$ & 437 \\
        \bottomrule
    \end{tabular}
    \caption{The popularity scores for the known CounterFact samples. The maximum popularity score measured was 1,176,235.}
    \label{tab:knowns-pop-score}
\end{table}

Using fact popularity, we also evaluate the known CounterFact samples through the lens of LM knowledge estimation. \Cref{tab:knowns-pop-score} lists the popularity scores distribution for the dataset. We find approximately 365 known CounterFact samples with popularity scores below 1000. These are unlikely to have been memorized by the model and are therefore unlikely to correspond to exact fact recall. Moreover, we find that around 50\% of these samples (172 samples) have been detected by our heuristics filters, indicating that the remaining samples may also contain surface level signals not detected by our filters. This supports our claim that popularity metadata can serve as a complement for separating exact fact recall samples from heuristics recall samples. 

\paragraph{Exact fact recall}
A total of 816 samples in CounterFact are found to have popularity scores above 1000, and are thus more likely to have been memorized by the model. We detect the potential usage of heuristics for 338 of these samples, meaning that approximately 478 samples in CounterFact may correspond to exact fact recall.

\subsection{Total effects}\label{app:knowns-te-details}
Apart from the analysis described above, we also scrutinize the known CounterFact samples with respect to the total effect of perturbing the subject. We measure the total effect on the probability of the output prediction. This provides an alternative way of checking for signs of lack of exact fact recall. The method was introduced by \citet{meng2022locating} and used to find model states important for the model prediction. By adding noise to the word embeddings corresponding to the subject of the query, the subject is perturbed. The idea is that the perturbation of the query makes the model incapable of performing the necessary recall of factual associations that resulted in the original prediction, thus lowering the model probability for the original prediction. We hypothesize that samples for which the added perturbation does not sufficiently lower the corresponding prediction probability are less likely to correspond to exact fact recall.

\paragraph{Method}
The total effect is measured as $\mathrm{TE}(o) = P_{\mathrm{clean}}(o)-P_{\mathrm{noised}}(o)$, where $P_{\mathrm{clean}}(o)$ denotes the probability of emitting token $o$ for a clean run and $P_{\mathrm{noised}}(o)$ denotes the probability of emitting token $o$ when the subject has been perturbed. For all our investigations, $o$ is given by the prediction corresponding to the query stored in the dataset. We note that negative total effects imply that the perturbation of the subject increased the probability of the original prediction and that low positive effects potentially indicate that perturbing the subject had a small effect on the model prediction.

Similarly to \citet{meng2022locating} we perturb the subject embeddings with noise $\epsilon \sim \mathrm{N}(0, \nu)$ where $\nu$ is set to be 3 times larger than the empirical standard deviation of all embeddings corresponding to the subjects of the dataset. We measure total effects for the known CounterFact samples as the average total effect of 10 runs with perturbed subjects.  

\paragraph{TE results}
For the 1209 known CounterFact samples we find 22 samples with negative total effects, i.e. perturbing the subject increased the prediction probability, of which 18 potentially correspond to prompt bias and 2 to name bias. Inspection of the samples marked for prompt bias reveal prompt patterns such as ``In [X], the language spoken is a mixture of'' where the corresponding prediction is ``English'' or ``German''. Another pattern we detect is ``[X] is affiliated with the religion of'' for which the prediction always is ``Islam''. We hypothesize that some prompts reveal the correct prediction even when the subject is occluded, resulting in negative TE values. 

\paragraph{Deeper study of TE results}
A deeper study of the TE values reveal an additional 37 samples for which the perturbation of the query subject decreased the original probability by less than 40\%. For some of these samples we identify queries that potentially reveal the correct prediction even when the subject is perturbed. Two identified samples are ``[X] professionally plays the sport of ice [hockey]'' or ``[X]'s expertise is in the field of quantum [physics]''. Prompt bias was detected for all of these queries. We measure a spearman correlation of -0.41 between normalized TE (\Cref{eq:norm-TE}) and the binary prompt bias metric over all known CounterFact samples. It is clear that the effect of perturbing the subject is smaller when the prediction is likely based on prompt bias, versus when it is not.

\begin{equation}\label{eq:norm-TE}
\mathrm{TE}_{\mathrm{norm}}(o) = \frac{P_{\mathrm{clean}}(o)-P_{\mathrm{noised}}(o)}{P_{\mathrm{clean}}(o)}
\end{equation}

\subsection{Negated queries}\label{app:knowns-not}
We identify a total of 8 problematic samples in the dataset that contain the word ``not'' in the query. Two examples are ``The language used by Louis Bonaparte is not the language of the [French]'' or ``The expertise of medical association is not in the field of [medicine]''. These samples are problematic as they are marked as correct since they contain the correct label, while they express the opposite of the fact represented by the data sample. This problem is a consequence of the sampling technique used by \citet{meng2022locating} in letting the LM generate a fluent continuation to a given query before making the prediction for the missing object. For the majority of the known CounterFact samples this leads to more fluent queries for which the LM might work better, but for some samples it results in reversed or revealing prompts.

\section{Normalization effects on causal tracing results}\label{app:normalization}

Since these results are dependent on the absolute values of the probability of the traced (predicted) token, we hypothesize that the result could be driven by a few high-probability samples and not representative of the low-probability\footnote{With \emph{probability}, we here refer to the probability corresponding to the clean run prediction.} strata of the data. 

To test this, we take inspiration from work by \citet{hase-does-localization-inform-editing} and compare the IE results to their normalized counterpart. We define the \emph{normalized indirect effect} as
\begin{equation}
\mathrm{NIE}_{h_i^{(l)}}(o) = \frac{P_{h_i^{(l)},\, \mathrm{patched}}(o)-P_{\mathrm{noised}}(o)}{|P_{\mathrm{clean}}(o)-P_{\mathrm{noised}}(o)|}
\end{equation}
where $P_{\mathrm{clean}}(o)-P_{\mathrm{noised}}(o)$ is the total effect (TE) defined as the difference between the clean and the noised runs. The normalized IE measures the percentage of recoverable probability that was recovered by patching state $h_i^{(l)}$.

For some samples, predominantly low-probability predictions, the division by the TE may result in unnatural $\mathrm{NIE}_{h_i^{(l)}}(o)$ values above 1 or below -1. The state patching should not be able to restore more than the clean run probability and we therefore cap the $\mathrm{NIE}_{h_i^{(l)}}(o)$ to a range of $[-1, 1]$.
With this approach, each sample is valued on the same scale. Plots for homogeneous datasets should therefore yield normalized CT results that are similar to their non-normalized counterparts.

\paragraph{Are aggregated CT results representative of each studied sample?}
The non-normalized results for combined samples seen in \Cref{fig:non_norm_CT_comb_lineplots} are dominated by the exact fact recall samples. The exact fact recall samples clearly lead to the decisive role conclusion and the same holds for the non-normalized results, even though subsets of the included data (heuristics recall and guesswork samples) do not lead to the same conclusion with as high certainty. 

For the normalized results we find that equal weights for all evaluated samples yield a slightly different pattern compared to the non-normalized results, with a weaker peak for the last subject token. 
We conclude that aggregations of CT results across multiple prediction scenarios are not representative of each studied sample. Also, comparisons between non-normalized and normalized results may reveal nonhomogeneous datasets with respect to prediction scenario. 

\section{Additional results from the CT sensitivity analysis}
This section contains additional results from the causal tracing analysis of \textsc{PrISM}.

\subsection{Llama 2 7B and 13B results}\label{app:CT-comb-llama2}
\begin{figure*}[h]
    \centering
    \begin{subfigure}[t]{0.366\textwidth}
        \centering
        \includegraphics[width=\linewidth]{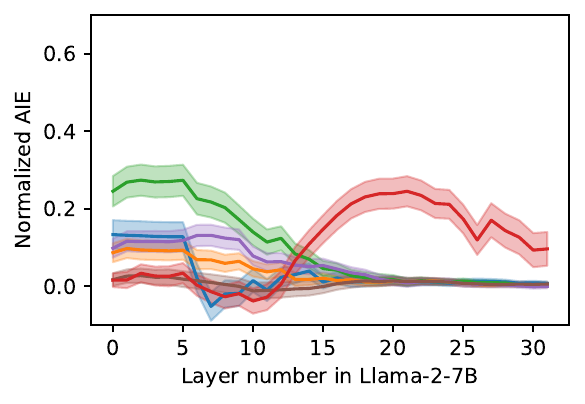}
        \vspace{-0.7cm}
        \caption{Generic language modeling samples.}
        \label{fig:CT_llama2_7B_generic_norm_lineplot1}
    \end{subfigure}%
    \begin{subfigure}[t]{0.317\textwidth}
        \centering
        \includegraphics[width=\linewidth,trim={38 0pt 0pt 0pt},clip]{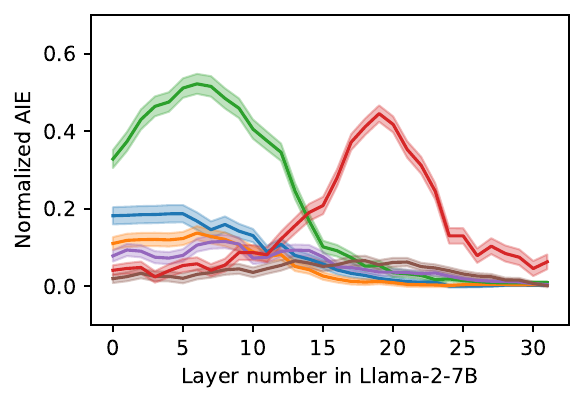}
        \vspace{-0.7cm}
        \caption{Guesswork samples.}
        \label{fig:CT_llama2_7B_guesswork_norm_lineplot}
    \end{subfigure}%
    \begin{subfigure}[t]{0.317\textwidth}
        \centering
        \vspace{-3.9cm}
        \hspace{-1.5cm}
        \includegraphics[width=0.7\linewidth,trim={2pt 7pt 2pt 2pt},clip]{figures/legend.pdf}
    \end{subfigure}
    \begin{subfigure}[t]{0.366\textwidth}
        \centering
        \includegraphics[width=\linewidth]{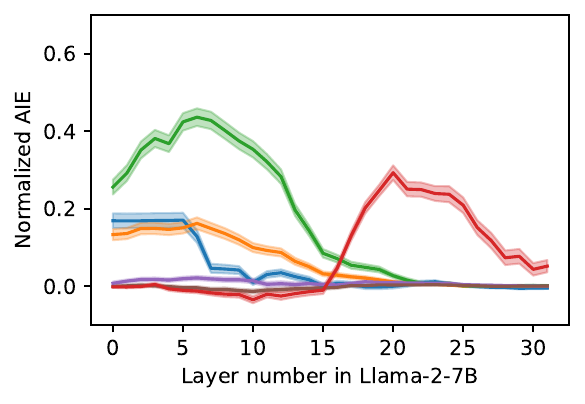}
        \vspace{-0.7cm}
        \caption{Heuristics recall samples.}
        \label{fig:CT_llama2_7B_biased_recall_norm_lineplot}
    \end{subfigure}%
    \begin{subfigure}[t]{0.317\textwidth}
        \centering
        \includegraphics[width=\linewidth,trim={38 0pt 0pt 0pt},clip]{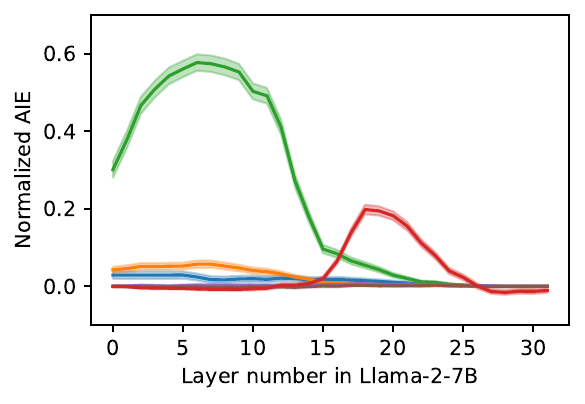}
        \vspace{-0.7cm}
        \caption{Exact fact recall samples.}
        \label{fig:CT_llama2_7B_exact_recall_norm_lineplot}
    \end{subfigure}%
    \begin{subfigure}[t]{0.317\textwidth}
        \centering
        \includegraphics[width=\linewidth,trim={38 0pt 0pt 0pt},clip]{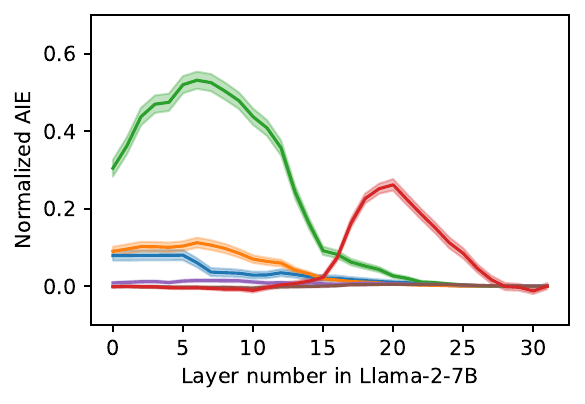}
        \vspace{-0.7cm}
        \caption{Combined samples.}
        \label{fig:CT_llama2_7B_comb_lineplot}
    \end{subfigure}
    \caption{CT results for \textsc{PrISM} Llama 2 7B data. 1000 samples for each scenario in isolation. As well as 1000 combined samples (330 exact fact recall, 340 heuristics recall, 330 guesswork). Shaded regions indicate 95\% confidence intervals.}
    \label{fig:CT_comb_lineplots_llama2_7B}
\end{figure*}

\begin{figure*}[h]
    \centering
    \begin{subfigure}[t]{0.366\textwidth}
        \centering
        \includegraphics[width=\linewidth]{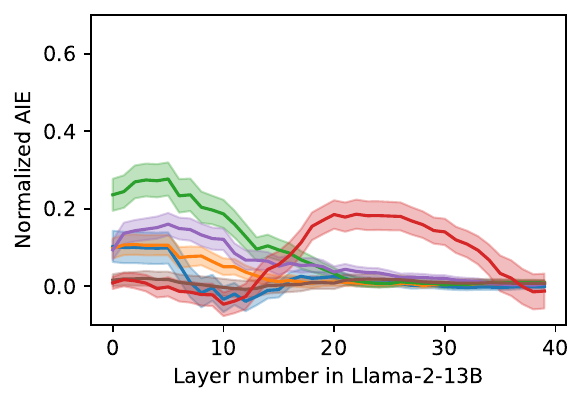}
        \vspace{-0.7cm}
        \caption{Generic language modeling samples.}
        \label{fig:CT_llama2_13B_generic_norm_lineplot1}
    \end{subfigure}%
    \begin{subfigure}[t]{0.317\textwidth}
        \centering
        \includegraphics[width=\linewidth,trim={38 0pt 0pt 0pt},clip]{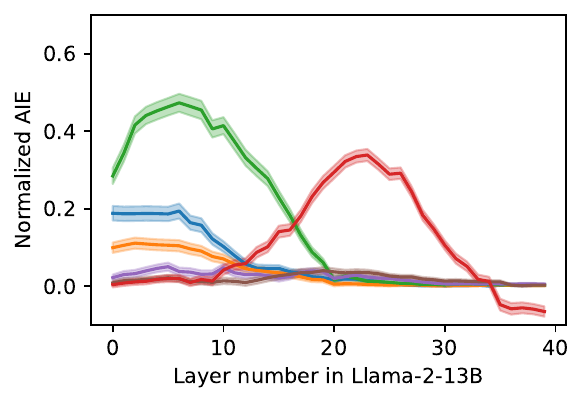}
        \vspace{-0.7cm}
        \caption{Guesswork samples.}
        \label{fig:CT_llama2_13B_guesswork_norm_lineplot}
    \end{subfigure}%
    \begin{subfigure}[t]{0.317\textwidth}
        \centering
        \vspace{-3.9cm}
        \hspace{-1.5cm}
        \includegraphics[width=0.7\linewidth,trim={2pt 7pt 2pt 2pt},clip]{figures/legend.pdf}
    \end{subfigure}
    \begin{subfigure}[t]{0.366\textwidth}
        \centering
        \includegraphics[width=\linewidth]{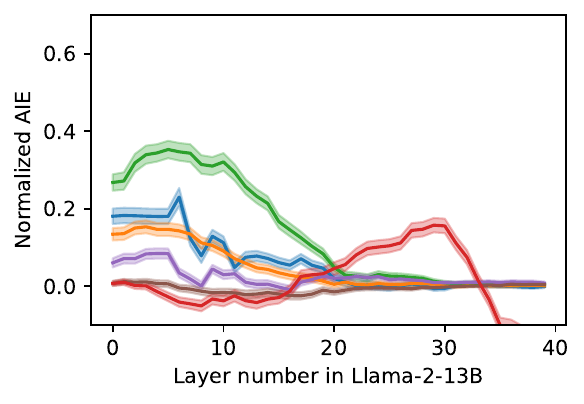}
        \vspace{-0.7cm}
        \caption{Heuristics recall samples.}
        \label{fig:CT_llama2_13B_biased_recall_norm_lineplot}
    \end{subfigure}%
    \begin{subfigure}[t]{0.317\textwidth}
        \centering
        \includegraphics[width=\linewidth,trim={38 0pt 0pt 0pt},clip]{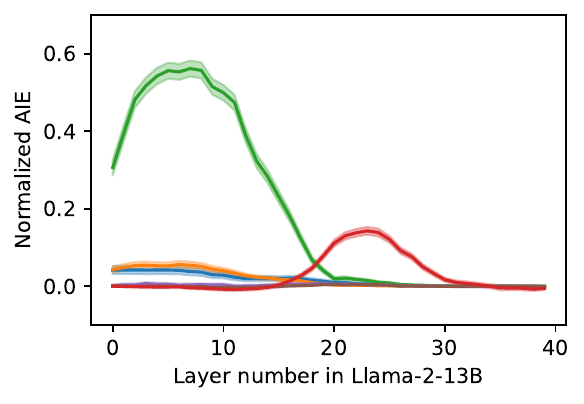}
        \vspace{-0.7cm}
        \caption{Exact fact recall samples.}
        \label{fig:CT_llama2_13B_exact_recall_norm_lineplot}
    \end{subfigure}%
    \begin{subfigure}[t]{0.317\textwidth}
        \centering
        \includegraphics[width=\linewidth,trim={38 0pt 0pt 0pt},clip]{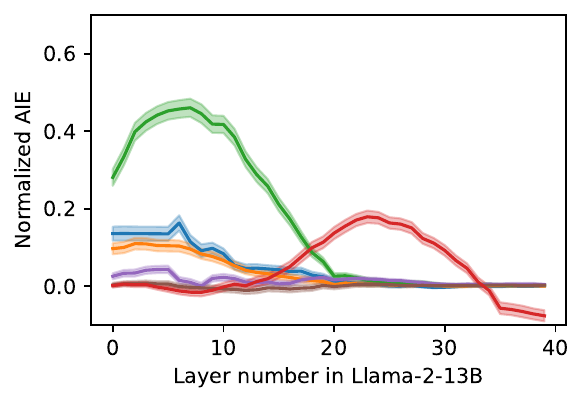}
        \vspace{-0.7cm}
        \caption{Combined samples.}
        \label{fig:CT_llama2_13B_comb_lineplot}
    \end{subfigure}
    \caption{CT results for \textsc{PrISM} Llama 2 13B data. 1000 samples for each scenario in isolation. As well as 1000 combined samples (330 exact fact recall, 340 heuristics recall, 330 guesswork). Shaded regions indicate 95\% confidence intervals.}
    \label{fig:CT_comb_lineplots_llama2_13B}
\end{figure*}

The results in \Cref{fig:CT_comb_lineplots_llama2_7B,fig:CT_comb_lineplots_llama2_13B} correspond to the results in \Cref{fig:CT_comb_lineplots_gpt2} but here for Llama 2 7B and 13B instead of GPT-2 XL. We find that the Llama results essentially support the same conclusions as the results for GPT-2 XL.

\subsection{Non-normalized results}
\begin{figure*}
    \centering
    \begin{subfigure}[t]{0.5\textwidth}
        \centering
        \includegraphics[width=\linewidth]{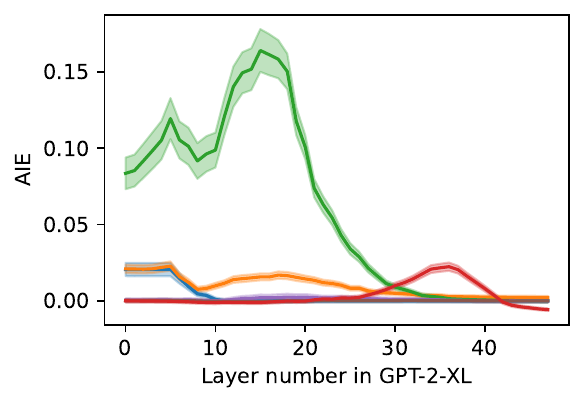}
        \vspace{-0.7cm}
        \caption{GPT-2 XL.}
        \label{fig:non_norm_CT_gpt2_xl}
    \end{subfigure}%
    \begin{subfigure}[t]{0.5\textwidth}
        \vspace{-5.3cm}
        \hspace{0cm}
        \includegraphics[width=0.5\linewidth,trim={2pt 7pt 2pt 2pt},clip]{figures/legend.pdf}
    \end{subfigure}
    \begin{subfigure}[t]{0.5\textwidth}
        \centering
        \includegraphics[width=\linewidth]{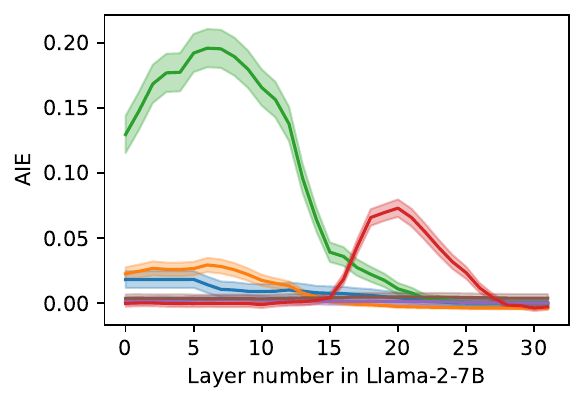}
        \vspace{-0.7cm}
        \caption{Llama 2 7B.}
        \label{fig:non_norm_CT_llama2_7B}
    \end{subfigure}%
    \begin{subfigure}[t]{0.5\textwidth}
        \centering
        \includegraphics[width=\linewidth]{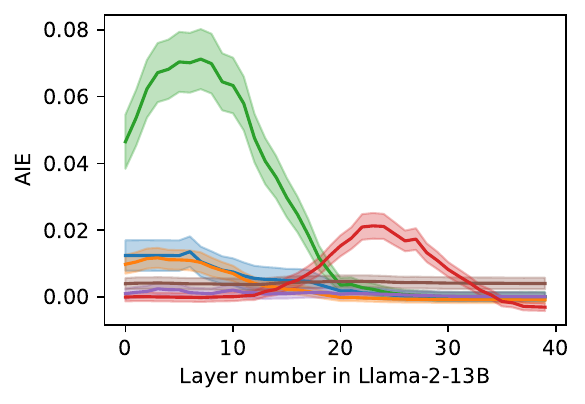}
        \vspace{-0.7cm}
        \caption{Llama 2 13B.}
        \label{fig:non_norm_CT_llama2_13B}
    \end{subfigure}
    \caption{Non-normalized CT results for the combined samples from \textsc{PrISM} for each of our studied models.}
    \label{fig:non_norm_CT_comb_lineplots}
\end{figure*}

To allow for comparisons with previous work that employed the CT method without normalization we present the non-normalized CT results for the combined samples in \Cref{fig:non_norm_CT_comb_lineplots}.

\subsection{Low-probability split}\label{app:CT-comb-low-probability}

\begin{figure*}[h]
    \centering
    \begin{subfigure}[t]{0.4\textwidth}
        \centering
        \includegraphics[width=\linewidth]{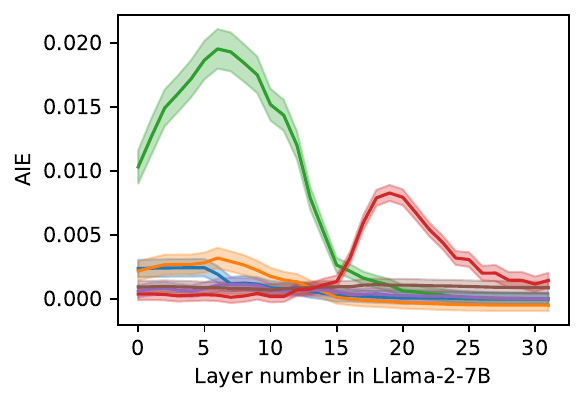}
        \vspace{-0.7cm}
        \caption{1000 combined samples.}
        \label{fig:CT_comb_lineplot_unconfident}
    \end{subfigure}%
    \begin{subfigure}[t]{0.395\textwidth}
        \centering
        \includegraphics[width=\linewidth]{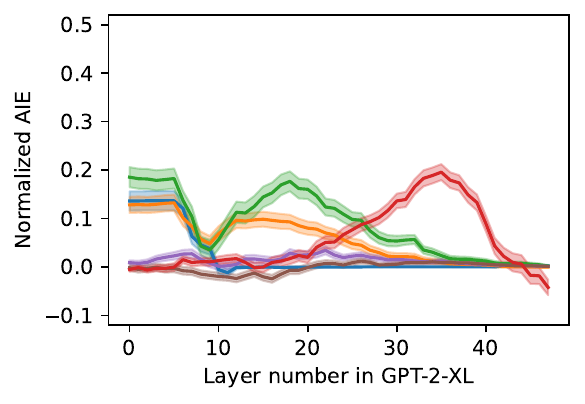}
        \vspace{-0.7cm}
        \caption{Normalized 1000 combined samples.}
        \label{fig:CT_comb_norm_lineplot_unconfident}
    \end{subfigure}%
    \begin{subfigure}[t]{0.205\textwidth}
        \centering
        \vspace{-4.2cm}
        \includegraphics[width=\linewidth,trim={2pt 4pt 2pt 2pt},clip]{figures/legend.pdf}
    \end{subfigure}
    \begin{subfigure}[t]{0.376\textwidth}
        \centering
        \includegraphics[width=\linewidth]{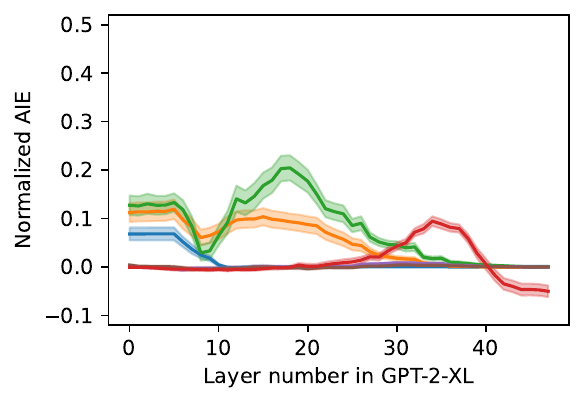}
        \vspace{-0.7cm}
        \caption{400 exact recall samples.}
        \label{fig:CT_exact_norm_lineplot_unconfident}
    \end{subfigure}%
    \begin{subfigure}[t]{0.312\textwidth}
        \centering
        \includegraphics[width=\linewidth,trim={48pt 0pt 0pt 0pt},clip]{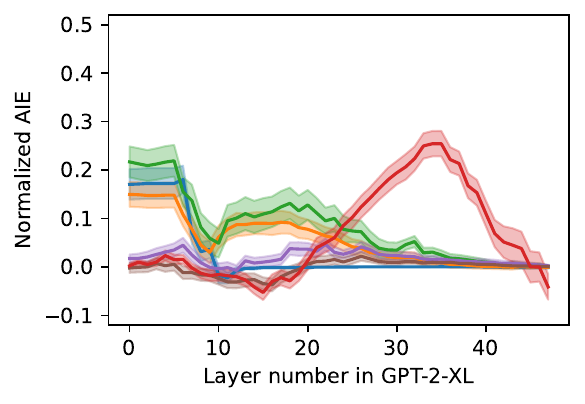}
        \vspace{-0.7cm}
        \caption{400 heuristics recall samples.}
        \label{fig:CT_biased_norm_lineplot_unconfident}
    \end{subfigure}%
    \begin{subfigure}[t]{0.312\textwidth}
        \centering
        \includegraphics[width=\linewidth,trim={48pt 0pt 0pt 0pt},clip]{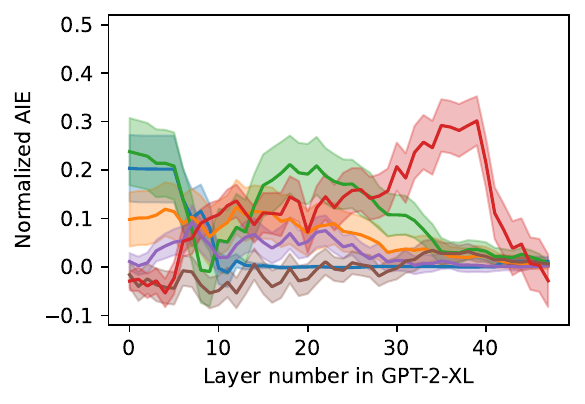}
        \vspace{-0.7cm}
        \caption{200 guesswork samples.}
        \label{fig:CT_guess_norm_lineplot_unconfident}
    \end{subfigure}
    \begin{subfigure}[t]{0.376\textwidth}
        \centering
        \includegraphics[width=\linewidth,trim={-10pt 0pt 0pt 0pt},clip]{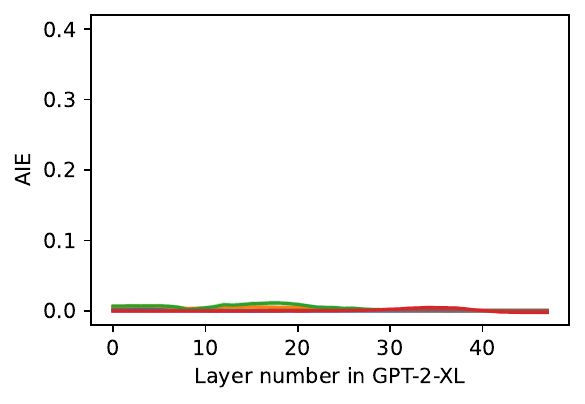}
        \vspace{-0.7cm}
        \caption{400 exact recall samples.}
        \label{fig:CT_exact_lineplot_unconfident}
    \end{subfigure}%
    \begin{subfigure}[t]{0.312\textwidth}
        \centering
        \includegraphics[width=\linewidth,trim={39pt 0pt 0pt 0pt},clip]{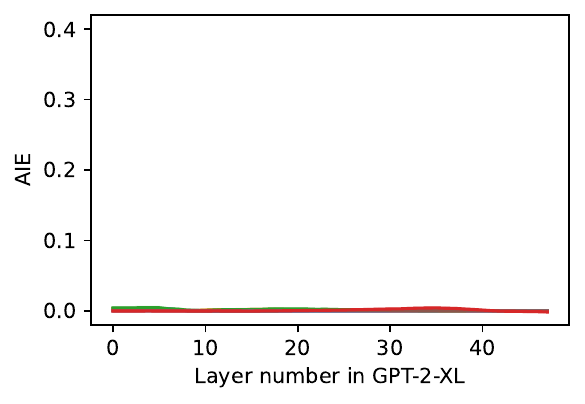}
        \vspace{-0.7cm}
        \caption{400 heuristics recall samples.}
        \label{fig:CT_biased_lineplot_unconfident}
    \end{subfigure}%
    \begin{subfigure}[t]{0.312\textwidth}
        \centering
        \includegraphics[width=\linewidth,trim={39pt 0pt 0pt 0pt},clip]{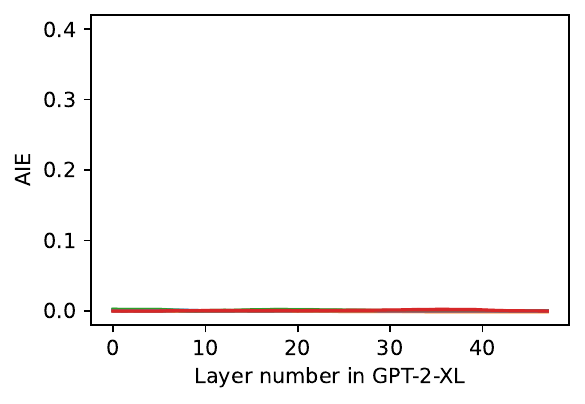}
        \vspace{-0.7cm}
        \caption{200 guesswork samples.}
        \label{fig:CT_guess_lineplot_unconfident}
    \end{subfigure}
    \caption{CT results on 1000 low-probability samples from \textsc{PrISM} of which 330 samples correspond to exact fact recall, 340 to heuristics recall and 330 to guesswork. These are the results for GPT-2 XL.}
    \label{fig:CT_comb_lineplots_gpt2_unconfident}
\end{figure*}

\begin{figure*}[h]
    \centering
    \begin{subfigure}[t]{0.4\textwidth}
        \centering
        \includegraphics[width=\linewidth]{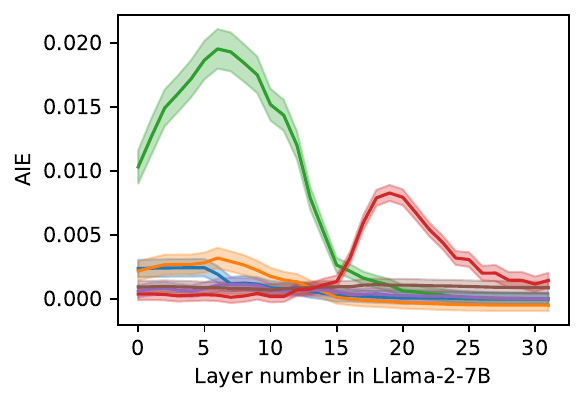}
        \vspace{-0.7cm}
        \caption{1000 combined samples.}
        \label{fig:CT_comb_lineplot_llama2_unconfident}
    \end{subfigure}%
    \begin{subfigure}[t]{0.395\textwidth}
        \centering
        \includegraphics[width=\linewidth]{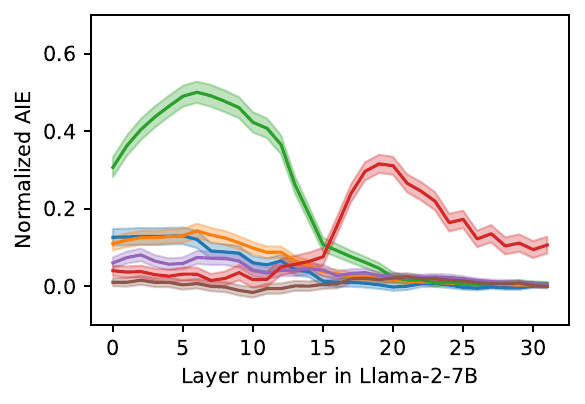}
        \vspace{-0.7cm}
        \caption{Normalized 1000 combined samples.}
        \label{fig:CT_comb_norm_lineplot_llama2_unconfident}
    \end{subfigure}%
    \begin{subfigure}[t]{0.205\textwidth}
        \centering
        \vspace{-4.2cm}
        \includegraphics[width=\linewidth,trim={2pt 4pt 2pt 2pt},clip]{figures/legend.pdf}
    \end{subfigure}
    \begin{subfigure}[t]{0.366\textwidth}
        \centering
        \includegraphics[width=\linewidth]{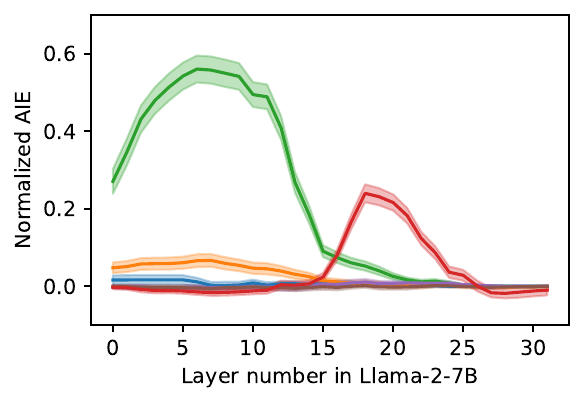}
        \vspace{-0.7cm}
        \caption{400 exact recall samples.}
        \label{fig:CT_exact_norm_lineplot_llama2_unconfident}
    \end{subfigure}%
    \begin{subfigure}[t]{0.317\textwidth}
        \centering
        \includegraphics[width=\linewidth,trim={38pt 0pt 0pt 0pt},clip]{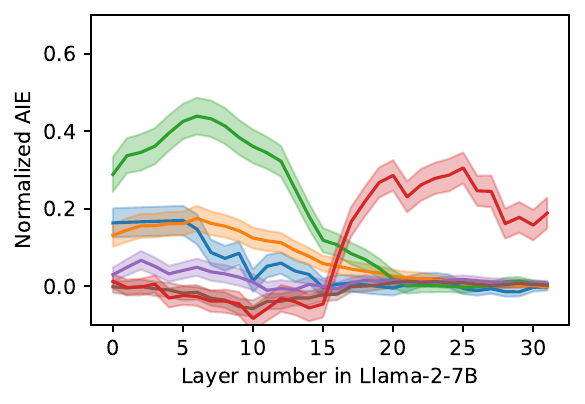}
        \vspace{-0.7cm}
        \caption{400 heuristics recall samples.}
        \label{fig:CT_biased_norm_lineplot_llama2_unconfident}
    \end{subfigure}%
    \begin{subfigure}[t]{0.317\textwidth}
        \centering
        \includegraphics[width=\linewidth,trim={38pt 0pt 0pt 0pt},clip]{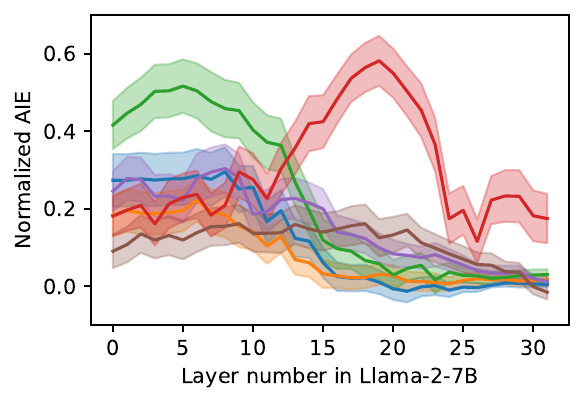}
        \vspace{-0.7cm}
        \caption{200 guesswork samples.}
        \label{fig:CT_guess_norm_lineplot_llama2_unconfident}
    \end{subfigure}
    \begin{subfigure}[t]{0.366\textwidth}
        \centering
        \includegraphics[width=\linewidth]{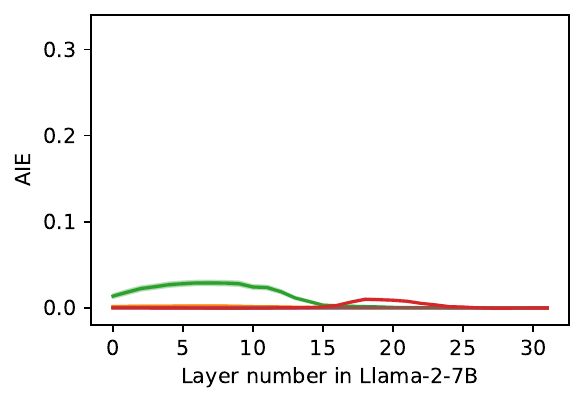}
        \vspace{-0.7cm}
        \caption{400 exact recall samples.}
        \label{fig:CT_exact_lineplot_llama2_unconfident}
    \end{subfigure}%
    \begin{subfigure}[t]{0.317\textwidth}
        \centering
        \includegraphics[width=\linewidth,trim={38pt 0pt 0pt 0pt},clip]{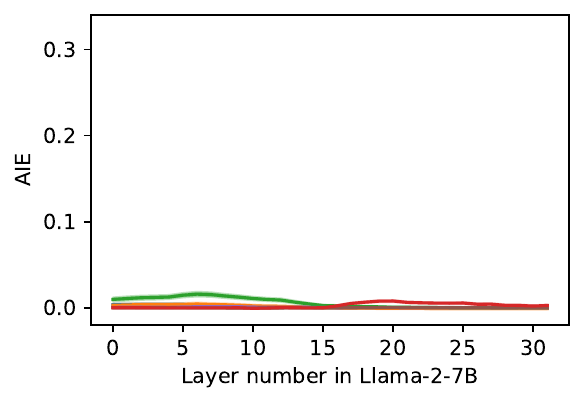}
        \vspace{-0.7cm}
        \caption{400 heuristics recall samples.}
        \label{fig:CT_biased_lineplot_llama2_unconfident}
    \end{subfigure}%
    \begin{subfigure}[t]{0.317\textwidth}
        \centering
        \includegraphics[width=\linewidth,trim={38pt 0pt 0pt 0pt},clip]{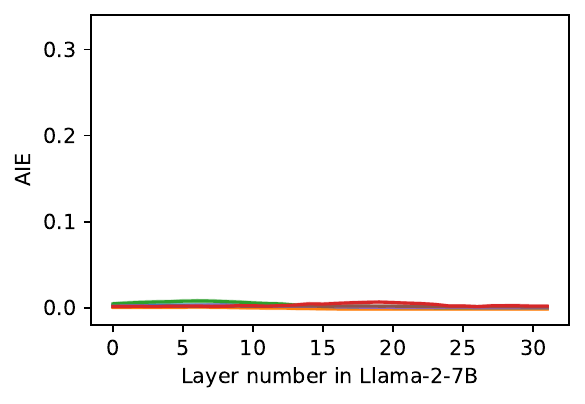}
        \vspace{-0.7cm}
        \caption{200 guesswork samples.}
        \label{fig:CT_guess_lineplot_llama2_unconfident}
    \end{subfigure}
    \caption{CT results on 1000 low-probability samples from \textsc{PrISM} of which 330 samples correspond to exact fact recall, 340 to heuristics recall and 330 to guesswork. These are the results for Llama 2 7B.}
    \label{fig:CT_comb_lineplots_llama2_unconfident}
\end{figure*}

The results in \Cref{fig:CT_comb_lineplots_gpt2,fig:CT_comb_lineplots_llama2_7B,fig:CT_comb_lineplots_llama2_13B} correspond to a sample of top-ranked prediction probabilities. The results in \Cref{fig:CT_comb_lineplots_gpt2_unconfident,fig:CT_comb_lineplots_llama2_unconfident} correspond to a sample of bottom-ranked prediction probabilities. We observe qualitative differences between the two figure pairs, where bottom-ranked probability set corresponds to larger effects for the last token state.

\subsection{Deeper study of heuristics recall}\label{app:biased-separated}
\begin{figure*}[h]
    \centering
    \hfill%
    \begin{subfigure}[t]{0.205\textwidth}
        \centering
        \includegraphics[width=\linewidth,trim={2pt 4pt 2pt 2pt},clip]{figures/legend.pdf}
    \end{subfigure}
    \begin{subfigure}[t]{0.5\textwidth}
        \centering
        \includegraphics[width=\linewidth]{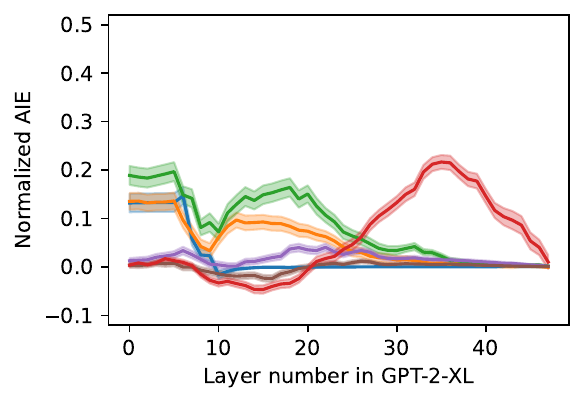}
        \vspace{-0.7cm}
        \caption{Prompt bias for GPT-2 XL.}
        \label{fig:CT_gpt2_prompt_bias_lineplot}
    \end{subfigure}%
    \begin{subfigure}[t]{0.5\textwidth}
        \centering
        \includegraphics[width=\linewidth]{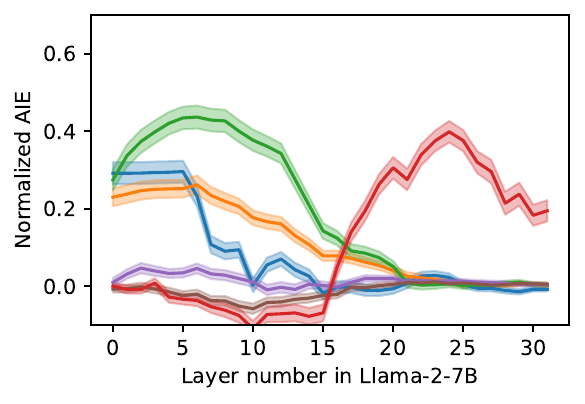}
        \vspace{-0.7cm}
        \caption{Prompt bias for Llama 2 7B.}
        \label{fig:CT_llama2_prompt_bias_lineplot}
    \end{subfigure}
    \begin{subfigure}[t]{0.5\textwidth}
        \centering
        \includegraphics[width=\linewidth]{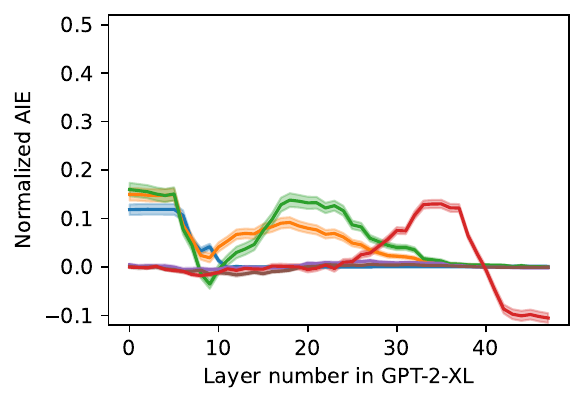}
        \vspace{-0.7cm}
        \caption{Person name bias for GPT-2 XL.}
        \label{fig:CT_gpt2_person_name_bias_lineplot}
    \end{subfigure}%
    \begin{subfigure}[t]{0.5\textwidth}
        \centering
        \includegraphics[width=\linewidth]{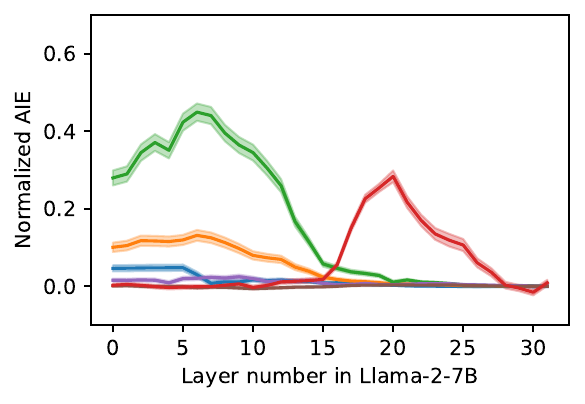}
        \vspace{-0.7cm}
        \caption{Person name bias for Llama 2 7B.}
        \label{fig:CT_llama2_person_name_bias_lineplot}
    \end{subfigure}
    \caption{Normalized CT results for sets of 1000 samples designed to exemplify each of the two main categories of the heuristics recall scenario.}
    \label{fig:CT_lineplots_bias}
\end{figure*}

We analyze the CT results of each of the main heuristics recall categories, prompt bias and person name bias, in separation for GPT-2 XL and Llama 2 7B. The corresponding results can be found in \Cref{fig:CT_lineplots_bias}. These results suggest a higher importance of the last token state, compared to the last subject token state, for the prompt bias subset compared to the person name bias subset. Potentially, it makes sense that prompt biased predictions that should be less sensitive to subject information attribute less importance to states corresponding to the subject.

\section{CT-based classifier for prediction scenario}\label{app:ct_class}

Our classifier is trained on 750 examples of each scenario (3000 data points in total) and performance is measured on a held out set of 1000 data points (150 of each scenario). We train a one-layer 50 -neuron neural network, with Adam optimizer, 0.0001 L2 regularization, 0.001 learning rate and a stopping tolerance of 1e-4. All models are trained until converging. \Cref{tab:NN_acc} shows overall performance and \Cref{tab:CM_GPT,,tab:CM_L7B,,tab:CM_L13B} show the respective confusion matrices.

\begin{table}[htb]
\centering
\begin{tabular}{ll}
\toprule
Data & Accuracy \\
\midrule
\textsc{PrISM} GPT2-XL & 0.73 \\
\textsc{PrISM} Llama 2 7B & 0.78 \\
\textsc{PrISM} Llama 3 13B & 0.74 \\
\bottomrule
\end{tabular}
\caption{\label{tab:NN_acc}Performance of a neural network classifier for predicting scenarios based on CT results.}
\end{table}

\begin{table}[htb]
\centering
\begin{tabular}{lllll}
\toprule
 & 0 & 1 & 2 & 3 \\
 \midrule
0 & 180 & 52 & 15 & 3 \\
1 & 28 & 184 & 29 & 9 \\
2 & 15 & 51 & 138 & 46 \\
3 & 2 & 11 & 10 & 227 \\
\bottomrule
\end{tabular}
\caption{\label{tab:CM_GPT}Confusion matrix for performance on \textsc{PrISM} GPT2-XL. Rows indicate true label, columns -- predictions. 0 = exact fact recall; 1 = heuristics; 2 = guesswork; 3 = generic LM}
\end{table}

\begin{table}[htb]
\centering
\begin{tabular}{lllll}
\toprule
 & 0 & 1 & 2 & 3 \\
\midrule
0 & 218 & 17 & 12 & 3 \\
1 & 23 & 165 & 30 & 32 \\
2 & 25 & 29 & 161 & 35 \\
3 & 1 & 8 & 7 & 234\\
\bottomrule
\end{tabular}
\caption{\label{tab:CM_L7B}Confusion matrix for performance on \textsc{PrISM} Llama 2 7B. Rows indicate true label, columns -- predictions. 0 = exact fact recall; 1 = heuristics; 2 = guesswork; 3 = generic LM}
\end{table}

\begin{table}[h]
\centering
\begin{tabular}{lllll}
\toprule
 & 0 & 1 & 2 & 3 \\
 \midrule
0 & 202 & 12 & 31 & 5 \\
1 & 15 & 167 & 35 & 33 \\
2 & 49 & 26 & 158 & 17 \\
3 & 5 & 19 & 13 & 213\\
\bottomrule
\end{tabular}
\caption{\label{tab:CM_L13B}Confusion matrix for performance on \textsc{PrISM} Llama 2 13B. Rows indicate true label, columns -- predictions. 0 = exact fact recall; 1 = heuristics; 2 = guesswork; 3 = generic LM}
\end{table}

\section{Additional results from the attribute extraction analysis for guesswork samples}\label{app:extraction-results}

\begin{figure*}[h]
    \centering
    \begin{subfigure}[b]{0.4\textwidth}
        \centering
        \includegraphics[width=\linewidth]{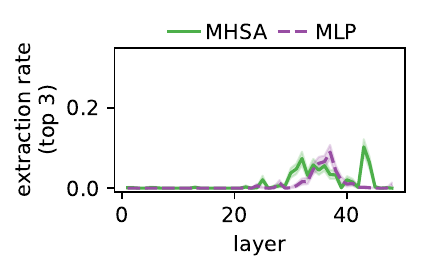}
        \vspace{-0.7cm}
        \caption{Top-3}
        \label{fig:extraction-guesswork-3}
    \end{subfigure}%
    \begin{subfigure}[b]{0.4\textwidth}
        \centering
        \includegraphics[width=\linewidth]{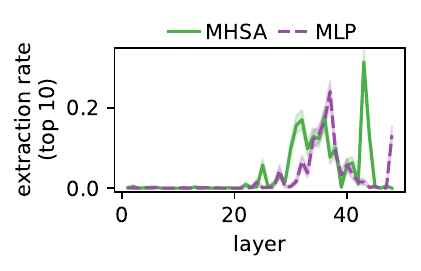}
        \vspace{-0.7cm}
        \caption{Top-10}
        \label{fig:extraction-guesswork-10}
    \end{subfigure}%
    \caption{Attribute extraction rates across layers measured for \textsc{PrISM} GPT-2 XL guesswork samples. MHSA and MLP indicate attribution extraction rates (top $k=3$ and top $k=10$) for multi-head self-attention states and multilayer perceptron states, respectively. Shaded regions indicate 95\% confidence intervals.}
    \label{fig:extraction-guesswork-extra}
\end{figure*}

Additional attribute extraction rate results for guesswork samples can be found in \Cref{fig:extraction-guesswork-extra}.

\section{Additional information flow results for heuristics recall samples}\label{app:dissection-results-heuristics}

\begin{figure*}[h]
    \centering
    \begin{subfigure}[b]{0.4\textwidth}
        \centering
        \includegraphics[width=\linewidth,trim={0pt 0pt 5pt 0pt},clip]{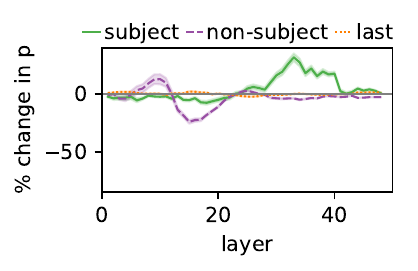}
        \vspace{-0.7cm}
        \caption{Prompt bias}
        \label{fig:flow-heuristics-prompt}
    \end{subfigure}%
    \begin{subfigure}[b]{0.4\textwidth}
        \centering
        \includegraphics[width=\linewidth,trim={0pt 0pt 5pt 0pt},clip]{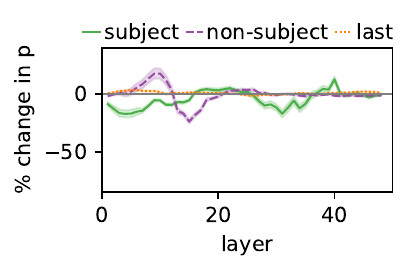}
        \vspace{-0.7cm}
        \caption{Person name bias}
        \label{fig:flow-heuristics-name}
    \end{subfigure}
    \caption{Relative change in the prediction probability when intervening on attention edges to the last position for window sizes of 9 layers in GPT-2 XL on prompt and person name bias samples. Shaded regions indicate 95\% confidence intervals.}
    \label{fig:flow-heuristics-extra}
\end{figure*}

\begin{figure*}[h]
    \centering
    \begin{subfigure}[b]{0.4\textwidth}
        \centering
        \includegraphics[width=\linewidth]{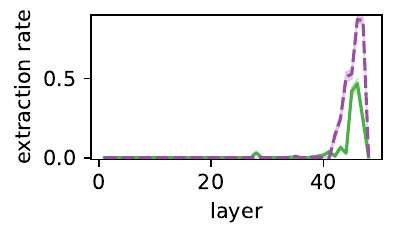}
        \vspace{-0.7cm}
        \caption{Prompt bias}
        \label{fig:extraction-heuristics-prompt}
    \end{subfigure}%
    \begin{subfigure}[b]{0.4\textwidth}
        \centering
        \includegraphics[width=\linewidth]{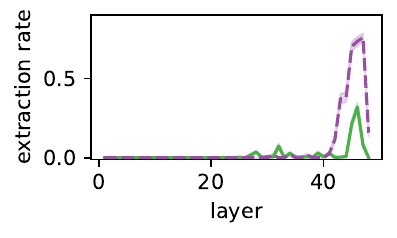}
        \vspace{-0.7cm}
        \caption{Person name bias}
        \label{fig:extraction-heuristics-name}
    \end{subfigure}
    \caption{Attribute extraction rates across layers measured for \textsc{PrISM} GPT-2 XL prompt and person name bias samples. MHSA and MLP indicate attribution extraction rates (top $k=1$) for multi-head self-attention states and multilayer perceptron states, respectively. Shaded regions indicate 95\% confidence intervals.}
    \label{fig:extraction-heuristics-extra}
\end{figure*}

Additional information flow and attribute extraction results for prompt and person name bias samples that make out the heuristics recall samples can be found in \Cref{fig:flow-heuristics-extra,fig:extraction-heuristics-extra}.

The results reveal similar extraction rate and information flow results for both heuristic types, while attention knockout to subject position states clearly \emph{increases} the prediction probability on prompt bias samples and slightly decreases the probability on person name bias samples. These results make sense, as prompt bias predictions should be independent from information about the subject, while a deeper analysis is necessary to better explain the low amplitudes measured in \Cref{fig:flow-heuristics,fig:flow-heuristics-extra,fig:extraction-heuristics-extra}.

\end{document}